\def\year{2018}\relax
\documentclass[letterpaper]{article}
\usepackage{aaai18}
\usepackage{times}
\usepackage{helvet}
\usepackage{courier}
\usepackage{url}
\usepackage{graphicx}

\usepackage{amsmath,amsfonts,amsthm,bm, amssymb}
\usepackage{subcaption}
\usepackage{graphicx}

\begin{document}
\title{IONet: Learning to Cure the Curse of Drift in Inertial Odometry}
\author{Changhao Chen, Xiaoxuan Lu, Andrew Markham, Niki Trigoni\\
Department of Computer Science, University of Oxford, United Kingdom\\
Email: \{firstname.lastname\}@cs.ox.ac.uk\\
}
\maketitle
\begin{abstract}
\begin{quote}
	Inertial sensors play a pivotal role in indoor localization, which in turn lays the foundation for pervasive personal applications. However, low-cost inertial sensors, as commonly found in smartphones, are plagued by bias and noise, which leads to unbounded growth in error when accelerations are double integrated to obtain displacement. Small errors in state estimation propagate to make odometry virtually unusable in a matter of seconds. We propose to break the cycle of continuous integration, and instead segment inertial data into independent windows. The challenge becomes estimating the latent states of each window, such as velocity and orientation, as these are not directly observable from sensor data. We demonstrate how to formulate this as an optimization problem, and show how deep recurrent neural networks can yield highly accurate trajectories, outperforming state-of-the-art shallow techniques, on a wide range of tests and attachments. In particular, we demonstrate that IONet can generalize to estimate odometry for non-periodic motion, such as a shopping trolley or baby-stroller, an extremely challenging task for existing techniques.
\end{quote}
\end{abstract}

\noindent Fast and accurate indoor localization is a fundamental need for many personal applications, including smart retail, public places navigation, human-robot interaction and augmented reality. One of the most promising approaches is to use inertial sensors to perform dead reckoning, which has attracted great attention from both academia and industry, because of its superior mobility and flexibility \cite{Lymberopoulos2015}.
    
    \begin{figure}
    	\centering
        \includegraphics[width=0.48\textwidth]{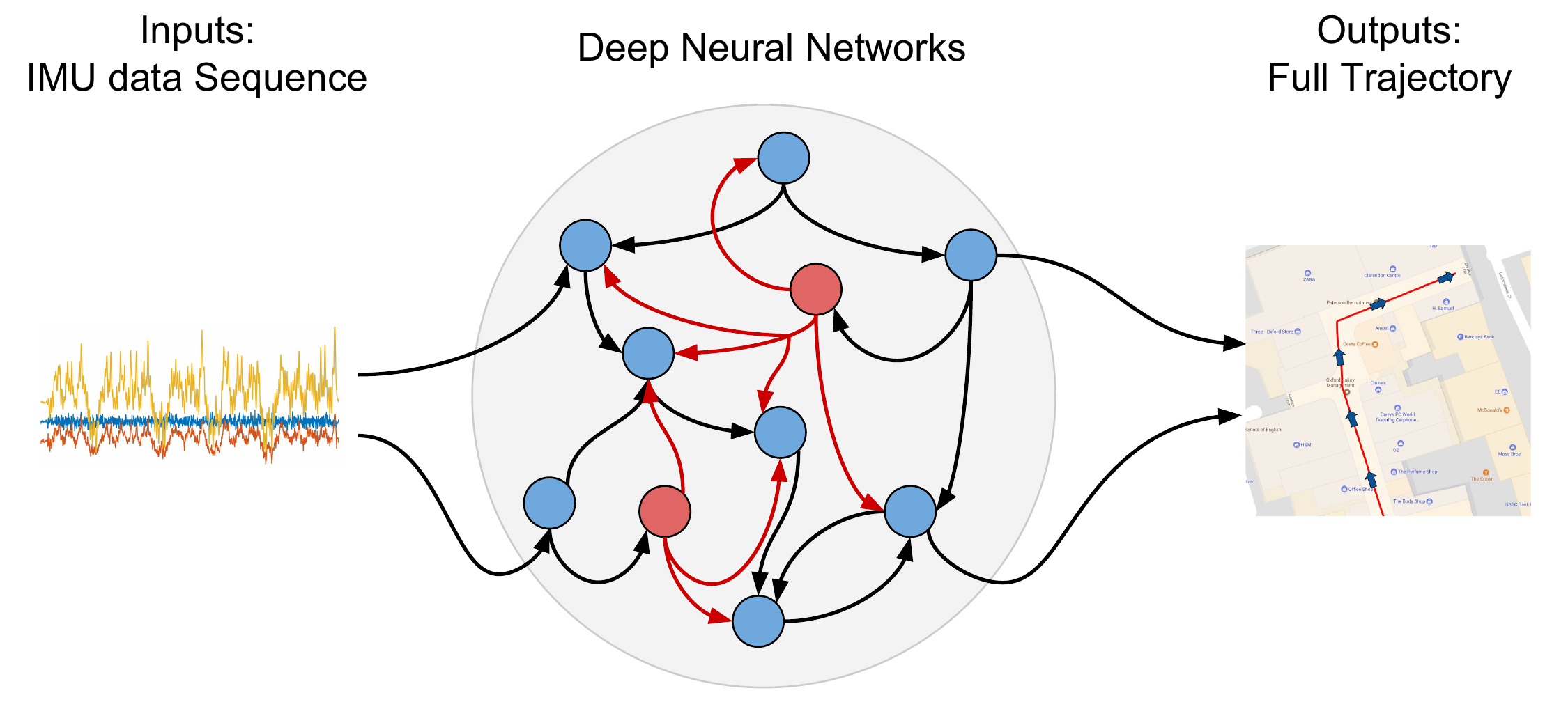}
        \caption{\label{fig:IONet} Overview of our proposed learning-based method}
    \end{figure} 
    
	Recent advances of MEMS (Micro-electro-mechanical systems) sensors have enabled inertial measurement units (IMUs) small and cheap enough to be deployed on smartphones. However, the low-cost inertial sensors on smartphones are plagued by high sensor noise, leading to unbounded system drifts. Based on Newtonian mechanics, traditional strapdown inertial navigation systems (SINS) integrate IMU measurements directly. They are hard to realize on accuracy-limited IMU due to exponential error propagation through integration. To address these problems, step-based pedestrian dead reckoning (PDR) has been proposed. This approach estimates trajectories by detecting steps, estimating step length and heading, and updating locations per step \cite{Li2012a}. Instead of double integrating accelerations into locations, a step length update mitigates exponential increasing drifts into linear increasing drifts. However, dynamic step estimation is heavily influenced by sensor noise, user's walking habits and phone attachment changes, causing unavoidable errors to the entire system \cite{Brajdic2013}. In some scenarios, no steps can be detected, for example, if a phone is placed on a baby stroller or shopping trolley, the assumption of periodicity, exploited by step-based PDR would break down. Therefore, the intrinsic problems of SINS and PDR prevent widespread use of inertial localization in daily life. The architecture of two existing methods is illustrated in Figure \ref{fig:existing_methods}.

    To cure the unavoidable `curse' of inertial system drifts, we break the cycle of continuous error propagation, and reformulate inertial tracking as a sequential learning problem. Instead of developing multiple modules for step-based PDR, our model can provide continuous trajectory for indoor users from raw data without the need of any hand-engineering, as shown in Figure \ref{fig:IONet}. Our contributions are three-fold:
    
\begin{itemize}
\item We cast the inertial tracking problem as a sequential learning approach by deriving a sequence-based physical model from Newtonian mechanics. 

\item We propose the first deep neural network (DNN) framework that learns location transforms in polar coordinates from raw IMU data, and constructs inertial odometry regardless of IMU attachment. 

\item We collected a large dataset for training and testing, and conducted extensive experiments across different attachments, users/devices and new environment, whose results outperform traditional SINS and PDR mechanisms. In addition, we demonstrate that our model can generalize to a more general motion without regular periodicity, e.g. trolley or other wheeled configurations. 
\end{itemize}

\section{Related Work}

	In this section, we provide a brief overview of some related work in inertial navigation systems, pedestrian dead reckoning (PDR) and sequential deep learning. 
    
    \textbf{Strapdown Inertial Navigation System}: strapdown inertial navigation systems (SINS) have been studied for decades \cite{Savage1998}. Previous inertial systems heavily relied on expensive, heavy, high-precision inertial measurement units, hence their main application had to be constrained on moving vehicles, such as automobiles, ships, aircraft, submarines and spacecraft. Recent advances of MEMS technology enable low-cost MEMS IMU to be deployed on robotics, UAV \cite{Bloesch2015}, and mobile devices \cite{Lymberopoulos2015}. However, restricted by size and cost, the accuracy of a MEMS IMU is extremely limited, and has to be integrated with other sensors, such as visual inertial odometry \cite{Leutenegger2015}. Another solution is to attach an IMU on the user's foot in order to take advantage of heel strikes for zero-velocity update to compensate system error drifts \cite{Skog2010}. These inconveniences prevent wide adoption of inertial solutions on consumer grade devices \cite{Harle2013}. 
    
    \textbf{Pedestrian Dead Reckoning}: Unlike SINS's open-loop integration of inertial sensors, PDR uses inertial measurements to detect steps, estimate stride length and heading via an empirical formula \cite{Shu2015a}. System errors still quickly accumulate, because of incorrect step displacement segmentation and inaccurate stride estimation. In addition, a large number of parameters have to be carefully tuned according to a user's walking habits. Recent research mainly focused on fusing PDR with external references, such as a floor plan \cite{Xiao2014a}, WiFi fingerprinting \cite{Hilsenbeck2014} or ambient magnetic fields \cite{Wang2016a}, still leaving fundamental problem of rapid drift unsolved. Compared with prior work, we abandon step-based approach and present a new general framework for inertial odometry. This allows us to handle  more general tracking problems, including trolley/wheeled configurations, which step-based PDR cannot address.
    
    \textbf{Sequential Deep Learning}: 
	Deep learning approaches have recently shown excellent performance in handling sequential data, such as speech recognition \cite{Graves2014}, machine translation \cite{Dai2015}, visual tracking \cite{Ondruska2016} and video description \cite{Donahue2015}. To the best of our knowledge, our IONet is the first neural network framework to achieve inertial odometry using inertial data only. Previous learning-based work has tackled localization problems, by employing visual odometry \cite{Zhou2017,Wang2017,Clark2017} and visual inertial odometry \cite{Clark2017a}. Some other work has concentrated on learning intuitive physics \cite{Hooman2017}, modeling state space models \cite{Karl2016}, and supervising neural networks via physics knowledge \cite{Stewart2017}. While most of them use visual observations, our work exploits real world sensor measurements to learn high-level motion trajectories.

\section{The Curse of Inertial Tracking}

The principles of inertial navigation are based on Newtonian mechanics. They allow tracking the position and orientation of an object in a navigation frame given an initial pose and measurements from accelerometers and gyroscopes.

	\begin{figure}
    	\centering
        \includegraphics[width=0.48\textwidth]{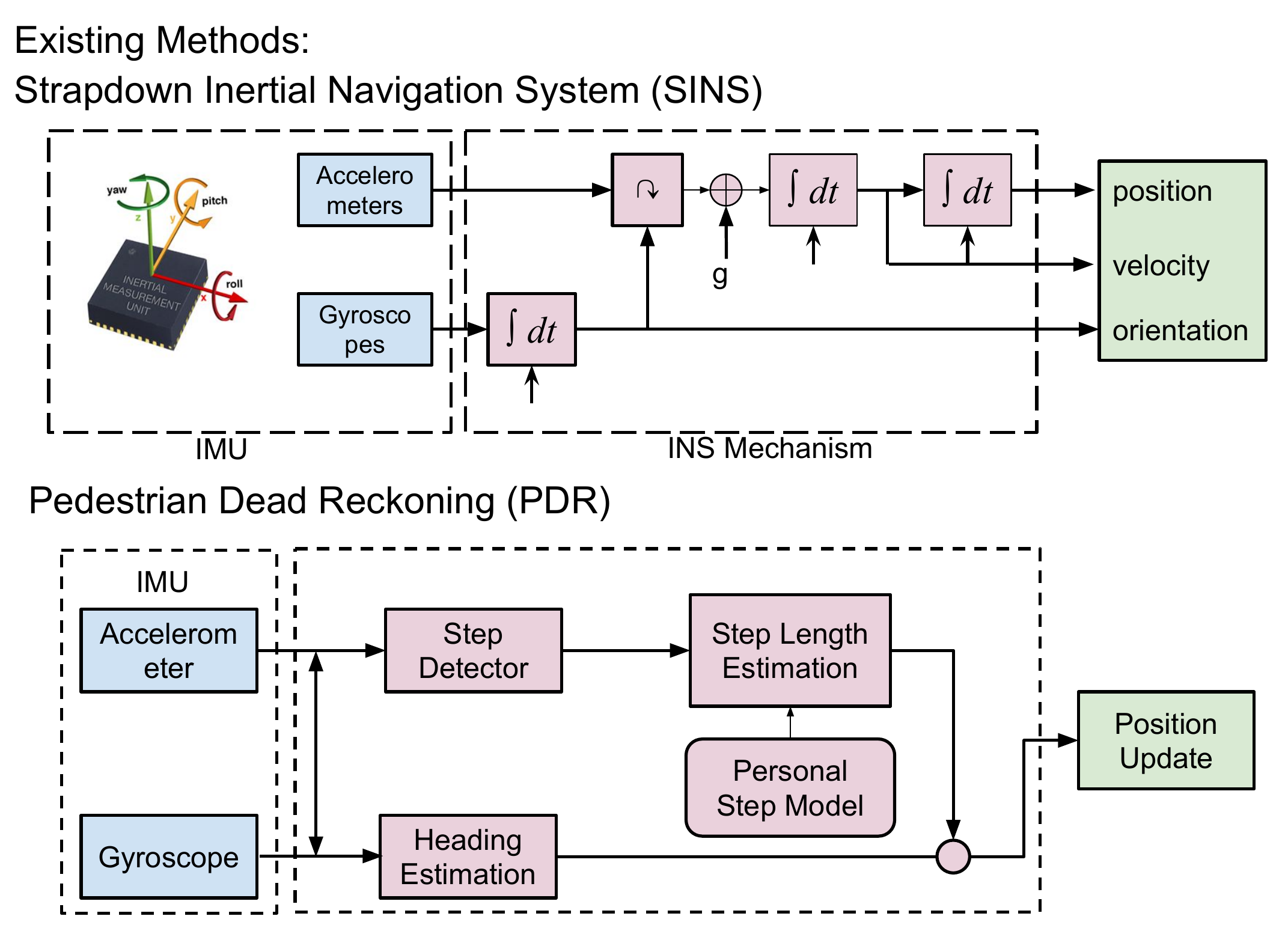}
        \caption{\label{fig:existing_methods} Architecture of existing methods: SINS and PDR}
    \end{figure} 

Figure \ref{fig:existing_methods} illustrates the basic mechanism of inertial navigation algorithms. The three-axis gyroscope measures angular velocities of the body frame with respect to the navigation frame, which are integrated into pose attitudes in Equations (\ref{eq:att_update1}-\ref{eq:att_update3}). To represent the orientation, the direction cosine $\mathbf{C}_b^n$ matrix is used to represent the transformation from the body (b) frame to the navigation (n) frame, and is updated with a relative rotation matrix $\mathbf{\Omega}(t)$. The 3-axis accelerometer measures proper acceleration vectors in the body frame, which are first transformed to the navigation frame and then integrated into velocity, discarding the contribution of gravity forces $\mathbf{g}$ in Equation (\ref{eq:vel_update}). The locations are updated by integrating velocity in Equation (\ref{eq:loc_update}). Equations (\ref{eq:att_update1}-\ref{eq:loc_update}) describe the attitude, velocity and location update at any time stamp $t$. In our application scenarios, the effects of earth rotation and the Coriolis accelerations are ignored. 

	Attitude Update:
	\begin{equation}	
    	\label{eq:att_update1}
    	\mathbf{C}_b^n(t)=\mathbf{C}_b^n(t-1)*\mathbf{\Omega}(t) \\      
    \end{equation}
    \begin{equation}
    	\label{eq:att_update2}
    	\mathbf{\sigma}=\mathbf{w}(t)dt
    \end{equation}
	\begin{equation}
    	\label{eq:att_update3}
		\mathbf{\Omega}(t)=\mathbf{C}_{b_t}^{b_{t-1}}=I+\frac{sin(\sigma)}{\sigma}[\mathbf{\sigma}\times]+\frac{1-cos(\sigma)}{\sigma^2}[\mathbf{\sigma}\times]^2
	\end{equation}
    
	Velocity Update:
	\begin{equation}
    	\label{eq:vel_update}
    	\mathbf{v}(t)=\mathbf{v}(t-1)+((\mathbf{C}_b^n(t-1))*\mathbf{a}(t)-\mathbf{g}_n)dt        
	\end{equation}

	Location Update:
	\begin{equation}
    	\label{eq:loc_update}
    	\mathbf{L}(t)=\mathbf{L}(t-1)+\mathbf{v}(t-1)dt    
    \end{equation}
where $\mathbf{a}$ and $\mathbf{w}$ are accelerations and angular velocities in body frame measured by IMU, $\mathbf{v}$ and $\mathbf{L}$ are velocities and locations in navigation frame, and $\mathbf{g}$ is gravity.
    
 	Under ideal condition, SINS sensors and algorithms can estimate system states for all future times. High-precision INS in military applications (aviation and marine/submarine) uses highly accurate and costly sensors to keep measurement errors very small. Meanwhile, they require a time-consuming system initialization including sensor calibration and orientation initialization. However, these requirements are inappropriate for pedestrian tracking. Realizing a SINS mechanism on low-cost MEMS IMU platform suffers from the following two problems: 
    
\begin{itemize}
\item The measurements from IMUs embedded in consumer phones are corrupted with various error sources, such as scale factor, axis misalignment, thermo-mechanical white noise and random walking noise \cite{NaserEl-SheimyHaiyingHou2008}. From attitude update to location update, the INS algorithm sees a triple integration from raw data to locations. Even a tiny noise will be highly exaggerated through this open-loop integration, causing the whole system to collapse within seconds.
\item A time-consuming initialization process is not suitable for everyday usage, especially for orientation initialization. Even small orientation errors would lead to the incorrect projection of the gravity vector. For example, a 1 degree attitude error will cause an additional 0.1712 $m/s^2$ acceleration on the horizontal plane, leading to 1.7 m/s velocity error and 8.56 m location error within 10 seconds.
\end{itemize}

Even if we view this physical model as a state-space-model, and apply a deep neural network directly to model inertial motion, the intrinsic problems would still `curse' the entire system.

\section{Tracking Down A Cure} 
	To address the problems of error propagation, our novel insight is to break the cycle of continuous integration, and segment inertial data into independent windows. This is analogous to resetting an integrator to prevent windup in classical control theory \cite{Hippe2006}.
    
    However, windowed inertial data is not independent, as Equations (\ref{eq:att_update1}-\ref{eq:loc_update}) clearly demonstrate. This is because key states (namely attitude, velocity and location) are \textit{unobservable} - they have to be derived from previous system states and inertial measurements, and propagated across time. Unfortunately, errors are also propagated across time, cursing inertial odometry. It is clearly impossible for windows to be truly independent. However, we can aim for pseudo-independence, where we estimate the \textit{change} in navigation state over each window. 
Our problem then becomes how to constrain or estimate these unobservable states over a window. Following this idea, we derive a novel sequence-based physical model from basic Newtonian Laws of Motion, and reformulate it into a learning model.
    
    The unobservable or latent system states of an inertial system consist of orientation $\mathbf{C}_b^n$, velocity $\mathbf{v}$ and position $\mathbf{L}$. In a traditional model, the transformation of system states could be expressed as a transfer function/state space model between two time frames in Equation (\ref{eq:transfer_function}), and the system states are directly coupled with each other. 
    \begin{equation}
    	\label{eq:transfer_function}
    	\begin{matrix}
        	[\mathbf{C}_b^n &  \mathbf{v} & \mathbf{L}]
        \end{matrix}_t = f(\begin{matrix} [\mathbf{C}_b^n &  \mathbf{v} & \mathbf{L}] \end{matrix}_{t-1}, \begin{matrix} [\mathbf{a} &  \mathbf{w}] \end{matrix}_t)
    \end{equation}
    
We first consider displacement. To separate the displacement of a window from the prior window, we compute the change in displacement $\Delta \mathbf{L}$ over an independent window of $n$ time samples, which is simply:
	\begin{equation}
    	\label{eq:displacement1}
		\Delta \mathbf{L} =\int_{t=0}^{n-1} \mathbf{v}(t) dt
	\end{equation}

We can separate this out into a contribution from the initial velocity state, and the accelerations in the navigation frame:
	\begin{equation}
    	\label{eq:displacement2}
		\Delta \mathbf{L} =n\mathbf{v}(0)dt+[(n-1)\mathbf{s}_1+(n-2)\mathbf{s}_2+\dotsi+\mathbf{s}_{n-1}]dt^2
	\end{equation}
where \begin{equation}
    	\mathbf{s}(t)=\mathbf{C}_b^n(t-1)\mathbf{a}(t)-\mathbf{g}
      \end{equation} is the acceleration in the navigation frame, comprising a dynamic part and a constant part due to gravity. 

Then, Equation (\ref{eq:displacement2}) is further formulated as:

	\begin{equation}
		\begin{split}
		\Delta \mathbf{L} &=n\mathbf{v}(0)dt+[(n-1)\mathbf{C}_b^n(0)*\mathbf{a}_1+(n-2)\mathbf{C}_b^n(0)\mathbf{\Omega}(1)\\
                  &*\mathbf{a}_2+\dotsi +\mathbf{C}_b^n(0)\displaystyle\prod_{i=1}^{n-2} \mathbf{\Omega}(i)*\mathbf{a}_{n-1}]dt^2-\frac{n(n-1)}{2}\mathbf{g}dt^2\\
		\end{split}
	\end{equation}
and simplified to become:
	\begin{equation}
		\begin{split}
			\Delta \mathbf{L} =n\mathbf{v}(0)dt+ \mathbf{C}_b^n(0) \mathbf{T} dt^2-\frac{n(n-1)}{2}\mathbf{g}dt^2 \\
		\end{split}
	\end{equation}
where \begin{equation}
    	\mathbf{T}=(n-1)\mathbf{a}_1+(n-2)\mathbf{\Omega}(1)\mathbf{a}_2+\dotsi+\displaystyle\prod_{i=1}^{n-2} \mathbf{\Omega}(i)\mathbf{a}_{n-1}
       \end{equation}

In our work, we consider the problem of indoor positioning i.e. tracking objects and people on a horizontal plane. This introduces a key observation: in the navigation frame, there is no long-term change in height\footnote{This assumption can be relaxed through the use of additional sensor modalities such as a barometer to detect changes in floor level due to stairs or elevator.}. 
The mean displacement in the z axis over a window is assumed to be zero and thus can be removed from the formulation. We can compute the absolute change in distance over a window as the L-2 norm i.e. $\Delta l=\|\Delta \mathbf{L} \|_2$, effectively decoupling the distance traveled from the orientation (e.g. heading angle) traveled, leading to:
	\begin{equation}
    	\begin{split}
        	&\Delta l=\| n\mathbf{v}(0)dt+ \mathbf{C}_b^n(0) \mathbf{T} dt^2-\frac{n(n-1)}{2}\mathbf{g}dt^2\|_2 \\
           &=\|\mathbf{C}_b^n(0)(n\mathbf{v}^b(0)dt+ \mathbf{T} dt^2-\frac{n(n-1)}{2}\mathbf{g}_0^b dt^2)\|_2
        \end{split}
    \end{equation}

Because the rotation matrix $\mathbf{C}_b^n(0)$ is an orthogonal matrix i.e. $\mathbf{C}_b^n(0)^T\mathbf{C}_b^n(0)=\mathbf{I}$, the initial unknown orientation has been successfully removed from, giving us: 
    \begin{equation}
    	\begin{split}
    		\Delta l =\|\Delta \mathbf{L} \|_2&=\| n\mathbf{v}^b(0)dt+ \mathbf{T} dt^2-\frac{n(n-1)}{2}\mathbf{g}_0^b dt^2\|_2 \\			
        \end{split}
    \end{equation}   
Hence, over a window, the horizontal distance traveled can be expressed as a function of the initial velocity, the gravity, and the linear and angular acceleration, all in the body frame:
	\begin{equation}
		\Delta l=f(\mathbf{v}^b(0), \mathbf{g}_0^b, \mathbf{a}_{1:n}, \mathbf{w}_{1:n})
	\end{equation}

To determine the change in the user's heading, we consider that a user's real accelerations and angular rates $(\mathbf{a}_{1:n},\mathbf{w}_{1:n})$ are also latent variables of IMU raw measurements $(\mathbf{\hat{a}}_{1:n},\mathbf{\hat{w}}_{1:n})$, and on the horizontal plane, only the heading attitude is essential in our system. From Equations (2-3) the change in the heading $\Delta \psi$ is expressed as a function of the raw data sequence. Therefore, we succeed in reformulating traditional model as a polar vector $(\Delta l, \Delta \psi)$ based model, which is only dependent on inertial sensor data, the initial velocity and gravity in the body frame:  
    \begin{equation}
    	\label{eq:polar_vector}
    		(\Delta l, \Delta \psi)=f_{\theta}(\mathbf{v}^b(0), \mathbf{g}_0^b, \mathbf{\hat{a}}_{1:n}, \mathbf{\hat{w}}_{1:n})
    \end{equation}
    
To derive a global location, the starting location $(x_0, y_0)$ and heading $\psi_0$ and the Cartesian projection of a number of windows can be written as, 
     \begin{equation}
    	\left\{
    	\begin{aligned}
    		x_n=x_0+\Delta l cos(\psi_0+\Delta \psi) \\
        	y_n=y_0+\Delta l sin(\psi_0+\Delta \psi)
        \end{aligned}
       \right.
    \end{equation}
    

Our task now becomes how to implicitly estimate this initial velocity and the gravity in body frame, by casting each window as a sequence learning problem.

This section serves as an overview showing the transition from the traditional model-based method to the proposed neural-network-based method. It takes the traditional state-space-model described in Equations (1-5), which converts raw data to poses in a step-by-step manner, to a formulation where a window of raw inertial data is processed in a batch to estimate a displacement and an angle change. Note that in both formulations, the final output depends on the initial attitude and velocity. As a result, in both cases, the curse of error accumulation won't be avoided if using the model-based integration approach. However, our sequence based formulation paves the way to our proposed neural network approach.

\section{Deep Neural Network Framework}
\begin{figure}
     	\centering
         \includegraphics[width=0.5\textwidth]{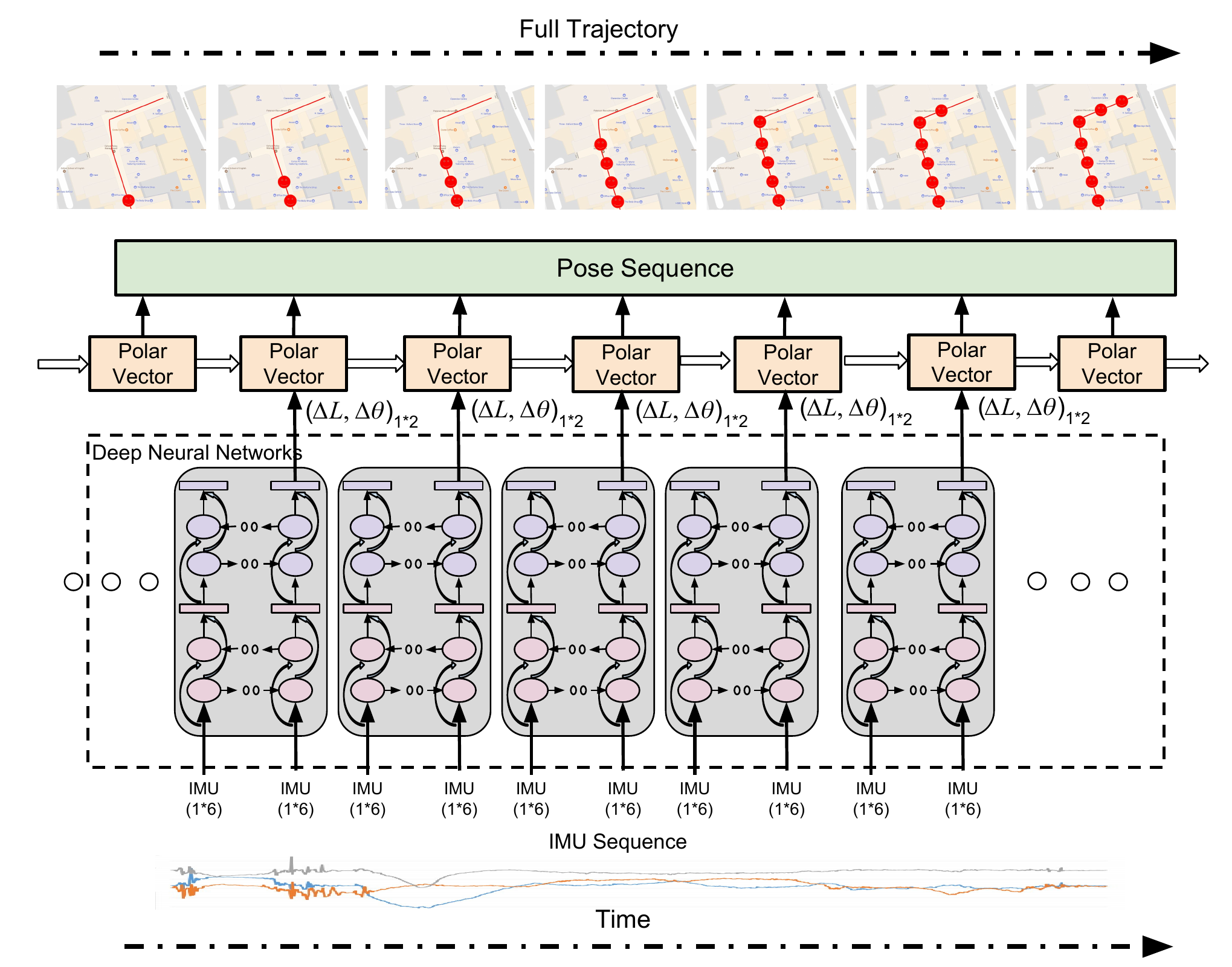}
         \caption{\label{fig:overview}Overview of IONet framework}
     \end{figure}

Estimating the initial velocity and the gravity in the body frame explicitly using traditional techniques is an extremely challenging problem. Rather than trying to determine the two terms, we instead treat Equation (\ref{eq:polar_vector}) as a sequence, where the inputs are the observed sensor data and the output is the polar vector. The unobservable terms simply become latent states of the estimation. Intuitively, the motivation for this relies on the regular and constrained nature of pedestrian motion. Over a window, which could be a few seconds long, a person walking at a certain rate induces a roughly sinusoidal acceleration pattern. The frequency of this sinusoid relates to the walking speed. In addition, biomechanical measurements of human motion show that as people walk faster, their strides lengthen \cite{Hausdorff2007}. Moreover, the gravity in body frame is related to the initial yaw and roll angle, determined by the attachment/placement of the device, which can be estimated from the raw data \cite{Xiao2015a}. In this paper, we propose the use of deep neural networks to learn the relationship between raw acceleration data and the polar delta vector, as illustrated in Figure \ref{fig:overview}.

Input data are independent windows of consecutive IMU measurements, strongly temporal dependent, representing body motions. To recover latent connections between motion characteristics and data features, a deep recurrent neural network (RNN) is capable of exploiting these temporal dependencies by maintaining hidden states over the duration of a window. Note however that latent states are not propagated between windows. Effectively, the neural network acts as a function $f_{\theta}$ that maps sensor measurements to polar displacement over a window
	\begin{equation}
		(\mathbf{a}, \mathbf{w})_{200*6} \xrightarrow{f_{\theta}} (\Delta l, \Delta \psi)_{1*2},
	\end{equation}
where a window length of 200 frames (2~s) is used here\footnote{We experimented with a window size of 50, 100, 200 and 400 frames, and selected 200 as an optimal parameter regarding the trade-off between accumulative location error and predicted loss.}.

	\begin{figure*}
    	\centering
        \begin{subfigure}[t]{0.3\textwidth}
        	\includegraphics[width=\textwidth]{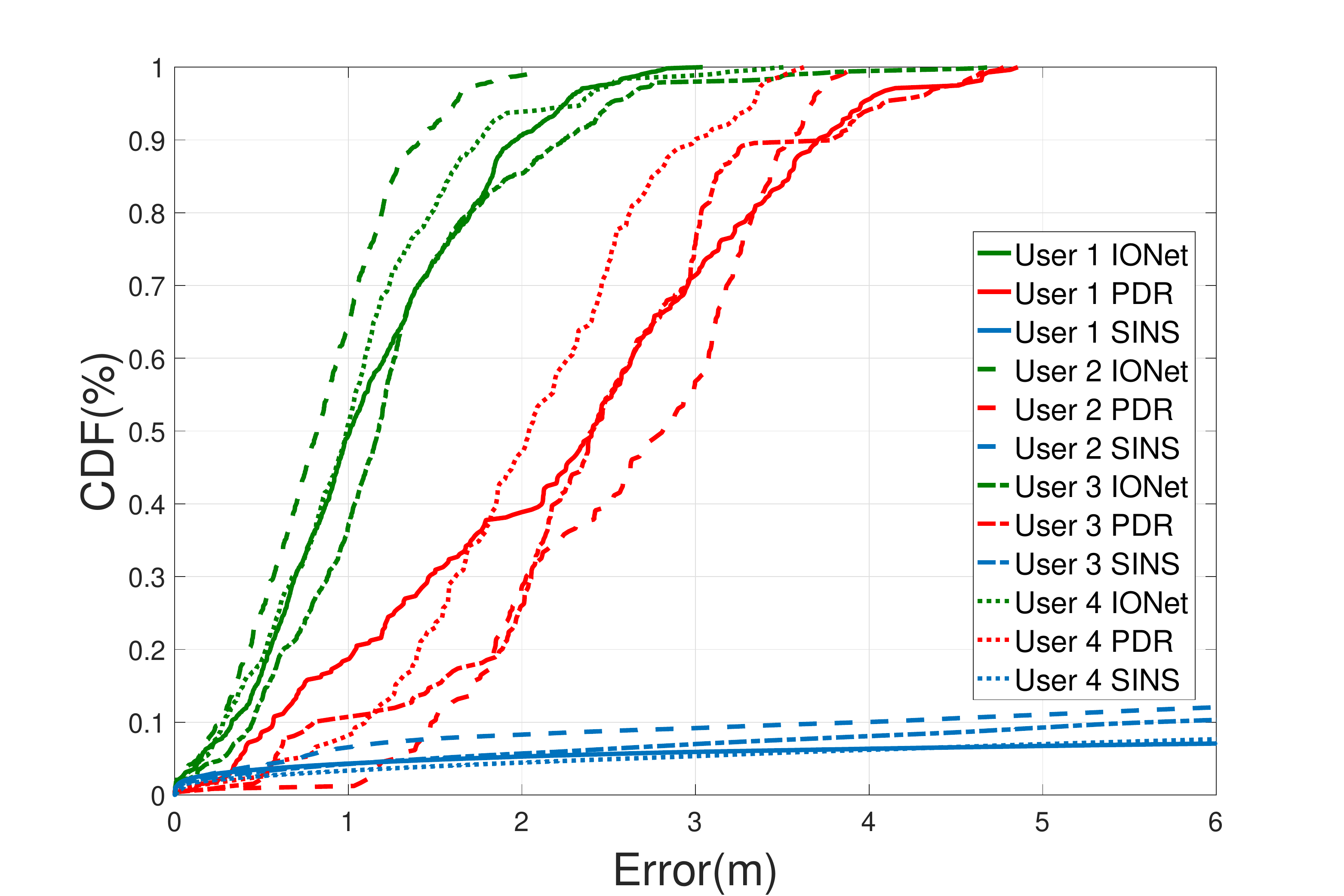}
        	\caption{\label{fig:hand_users} Handheld}
        \end{subfigure}
        \begin{subfigure}[t]{0.3\textwidth}
        	\includegraphics[width=\textwidth]{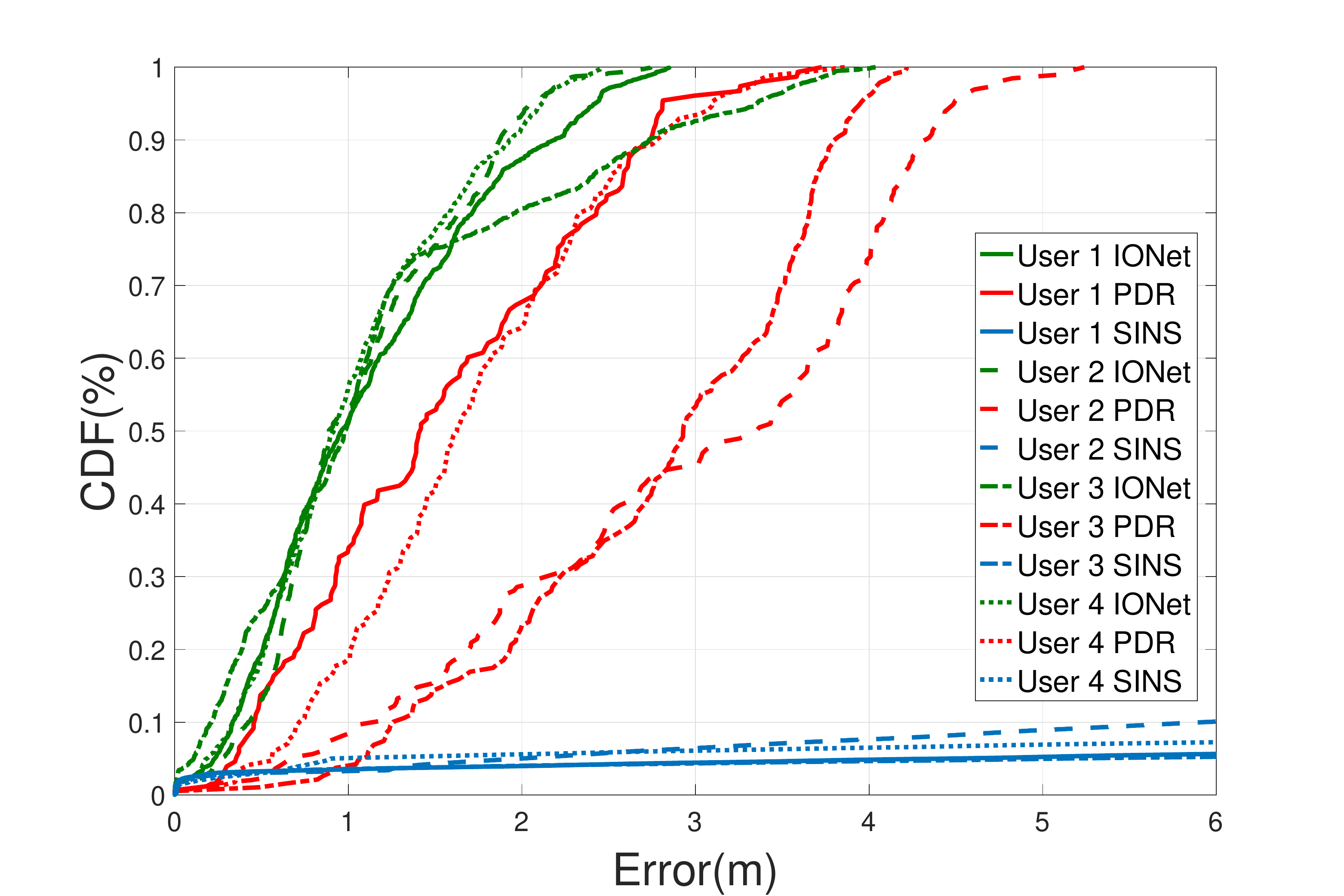}
        	\caption{\label{fig:pocket_users} In Pocket}
        \end{subfigure}
        \begin{subfigure}[t]{0.3\textwidth}
        	\includegraphics[width=\textwidth]{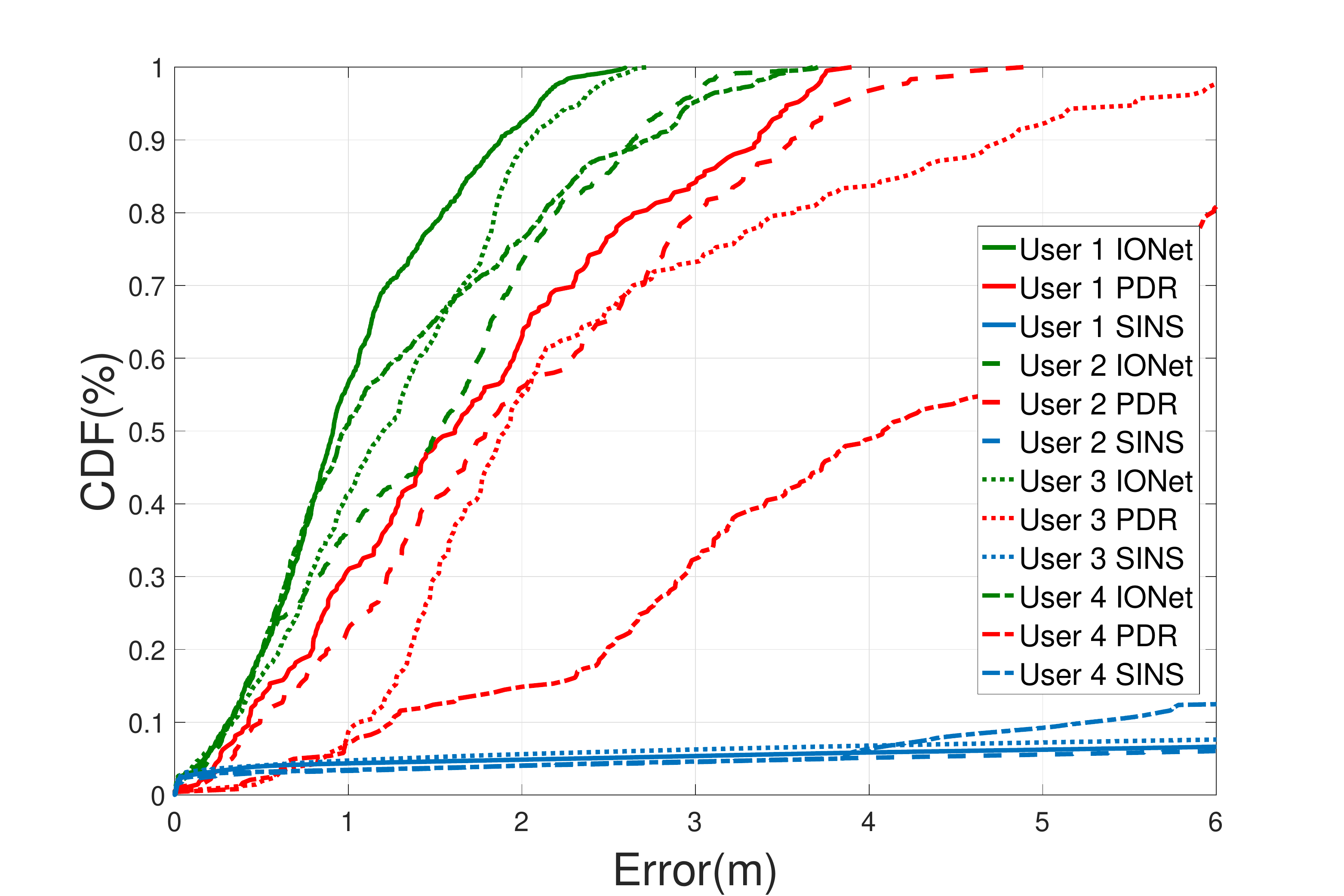}
        	\caption{\label{fig:handbag_users} In Handbag}
        \end{subfigure}
        \caption{\label{fig:multi_users} Performance in experiments involving different users.}
    \end{figure*}

	\begin{figure*}
    	\centering
        \begin{subfigure}[t]{0.3\textwidth}
        	\includegraphics[width=\textwidth]{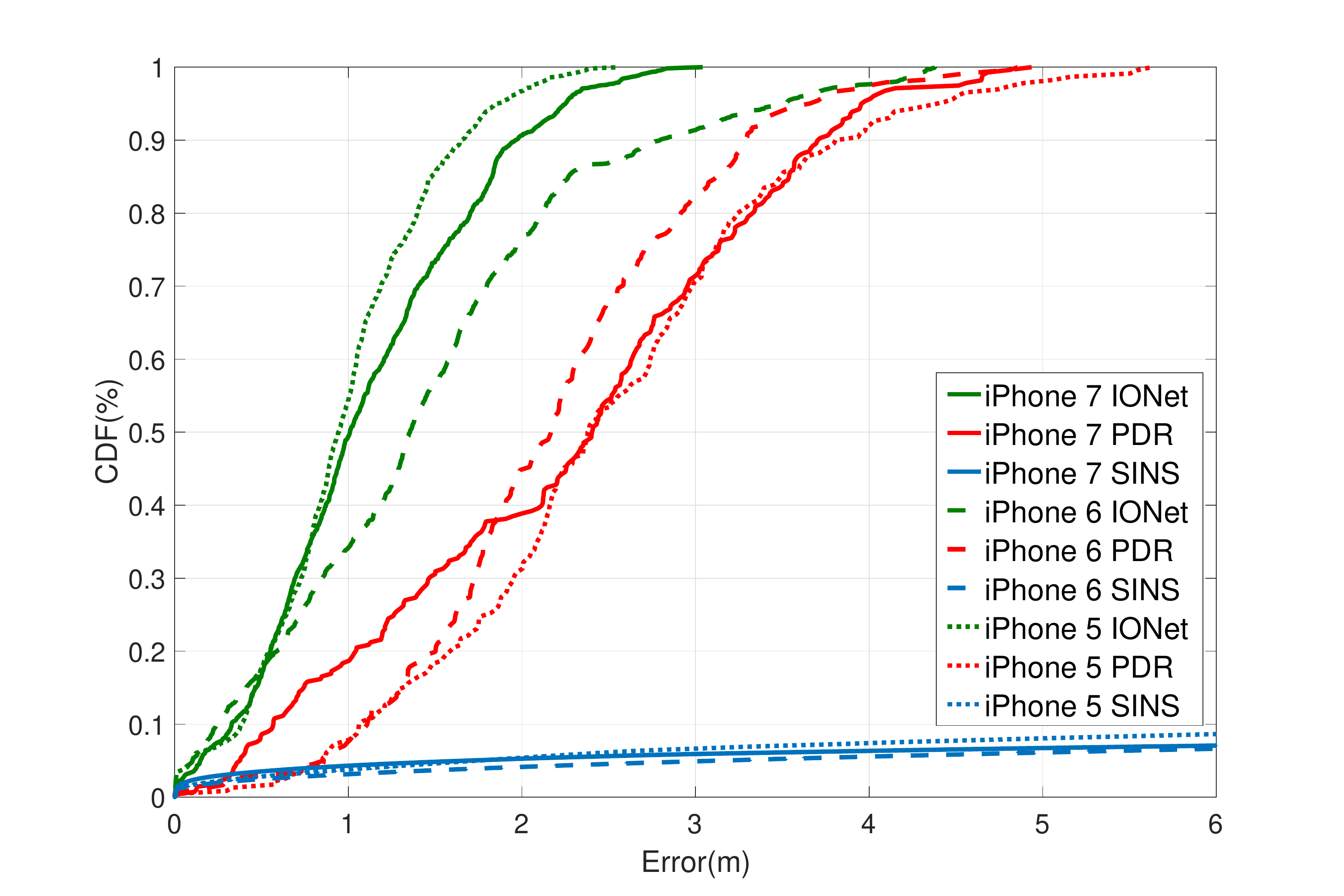}
        	\caption{\label{fig:hand_phones} Handheld}
        \end{subfigure}
        \begin{subfigure}[t]{0.3\textwidth}
        	\includegraphics[width=\textwidth]{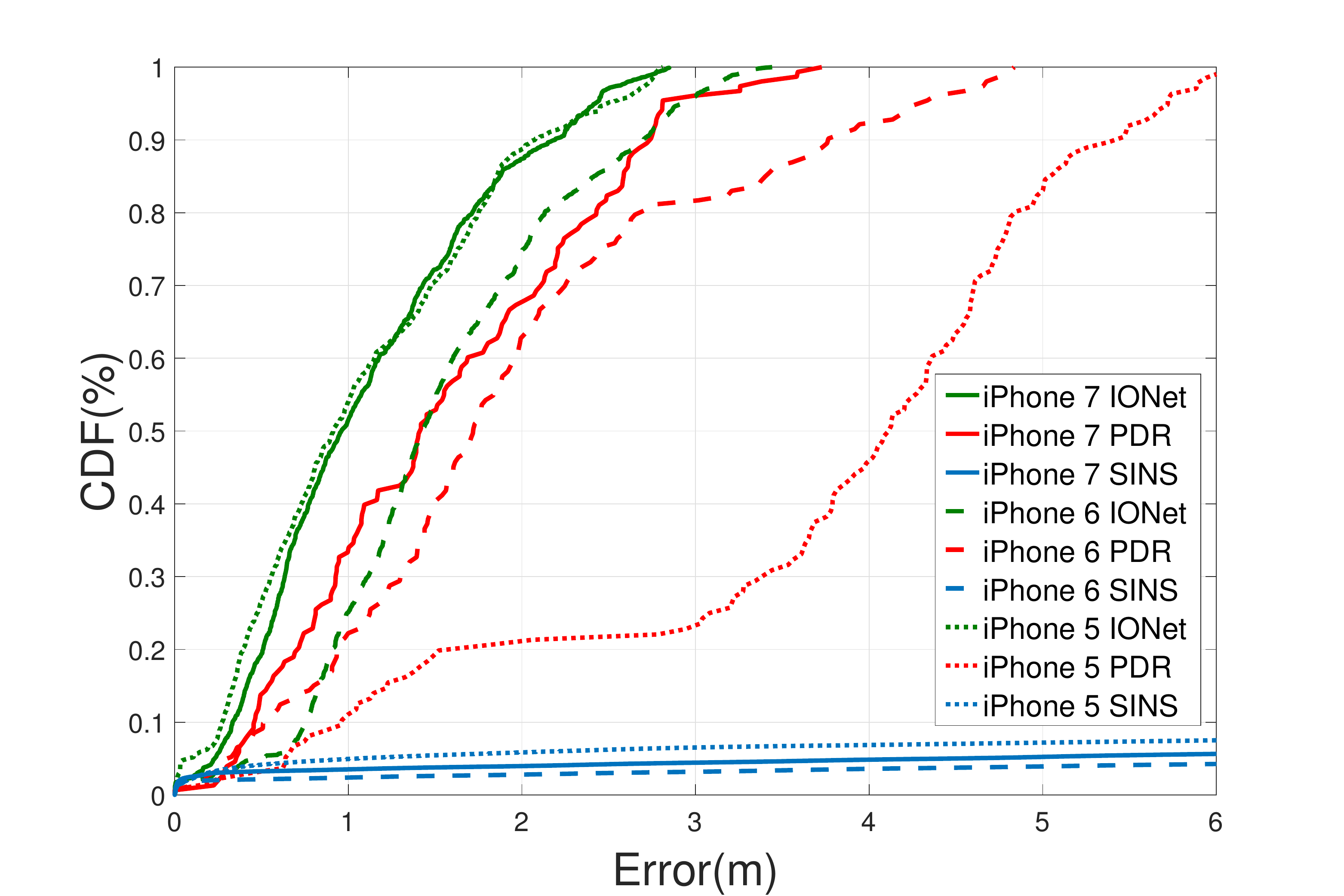}
        	\caption{\label{fig:pocket_phones} In Pocket}
        \end{subfigure}
        \begin{subfigure}[t]{0.3\textwidth}
        	\includegraphics[width=\textwidth]{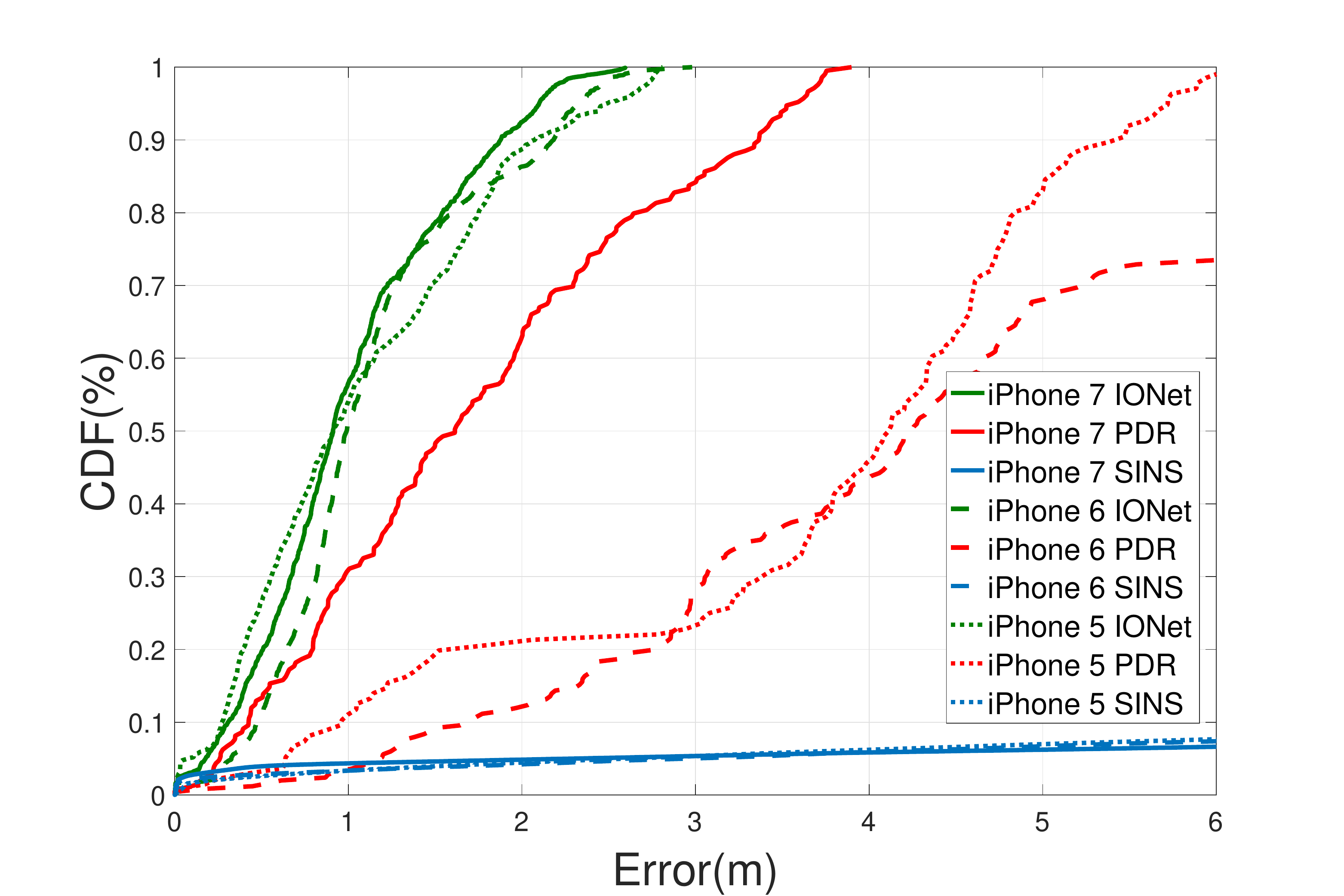}
        	\caption{\label{fig:handbag_phones} Handbag}
        \end{subfigure}
        \caption{\label{fig:multi_phones} Performance in experiments involving different devices.}
    \end{figure*}

	\begin{figure*}
    	\centering
        \begin{subfigure}[t]{0.3\textwidth}
        	\includegraphics[width=\textwidth]{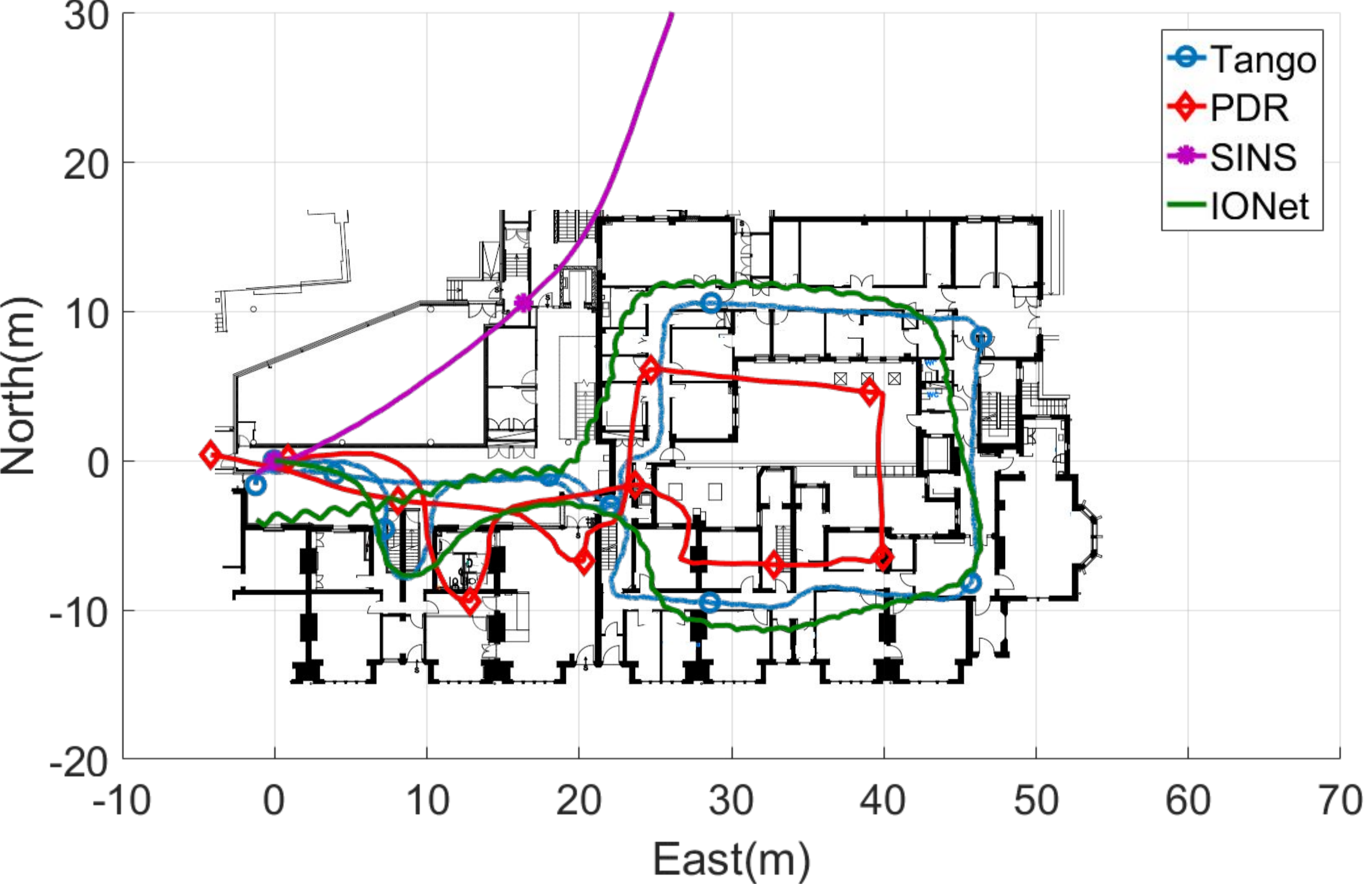}
        	\caption{\label{fig:floor1_hand_phones} Handheld}
        \end{subfigure}
        \begin{subfigure}[t]{0.3\textwidth}
        	\includegraphics[width=\textwidth]{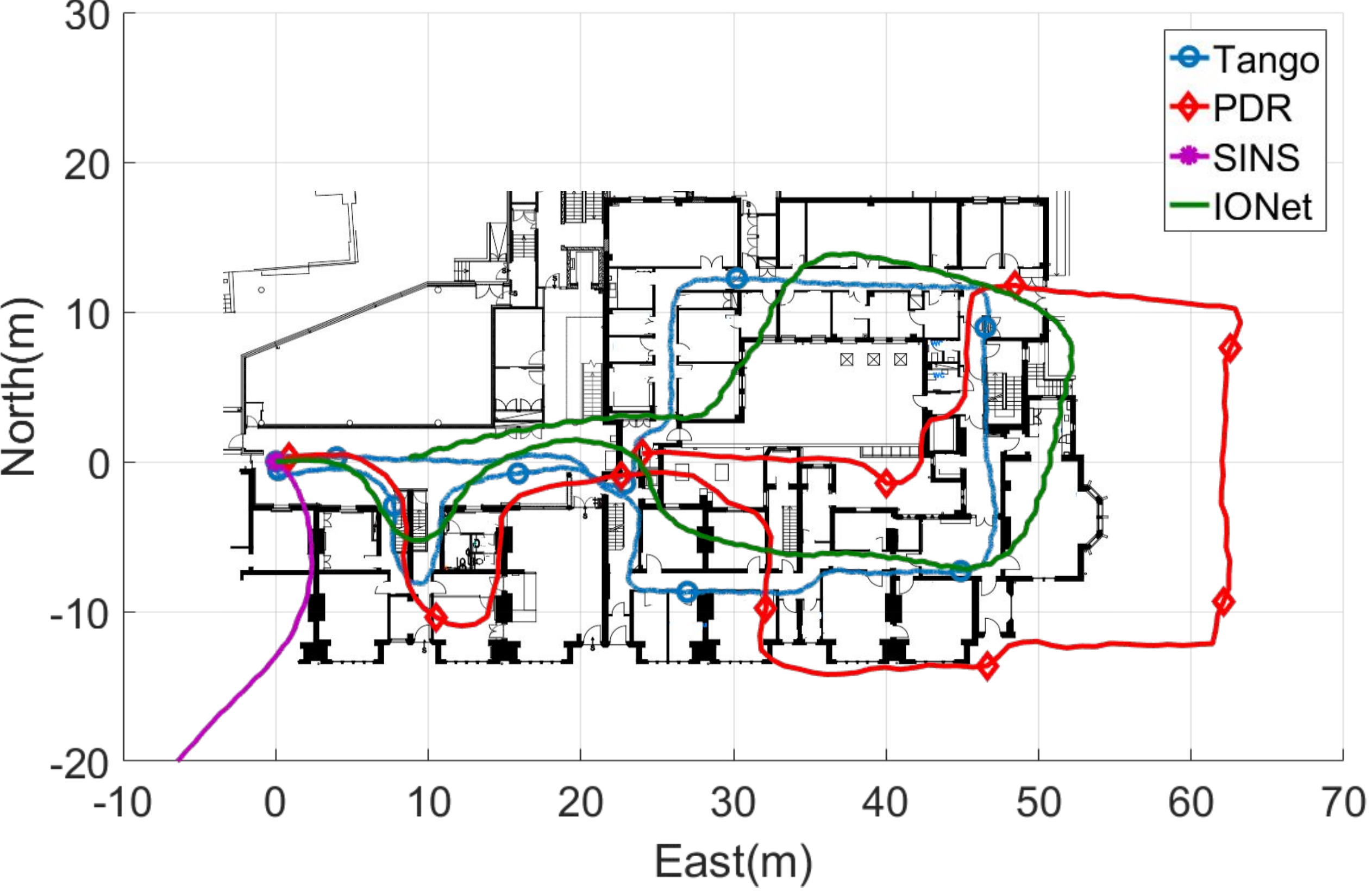}
        	\caption{\label{fig:floor1_pocket_phones} In Pocket}
        \end{subfigure}
        \begin{subfigure}[t]{0.3\textwidth}
        	\includegraphics[width=\textwidth]{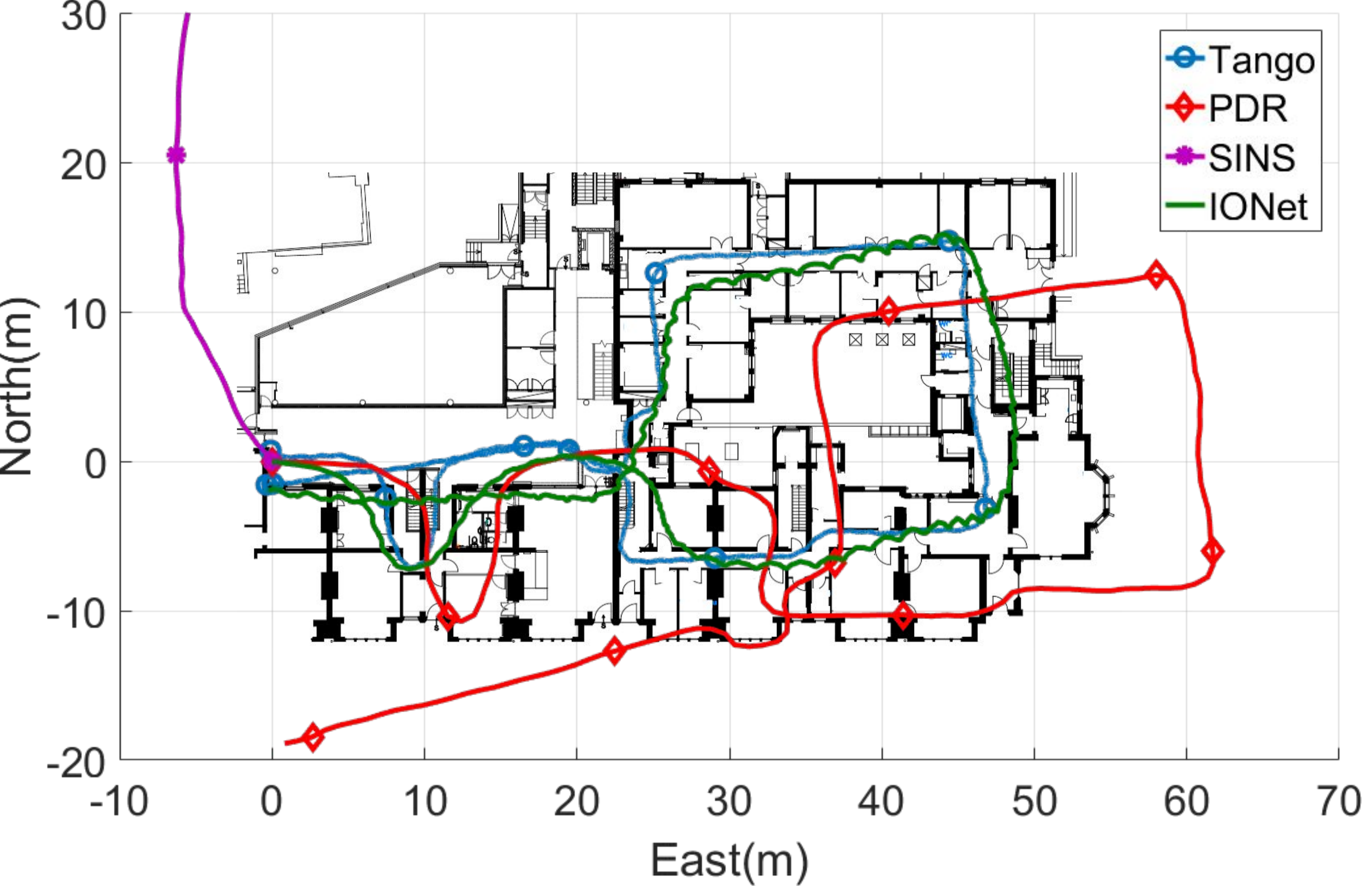}
        	\caption{\label{fig:floor1_handbag_phones} In Handbag}
        \end{subfigure}
        \caption{\label{fig:floor1} Trajectories on Floor A}
    \end{figure*}
    
    \begin{figure*}
    	\centering
        \begin{subfigure}[t]{0.3\textwidth}
        	\includegraphics[width=\textwidth]{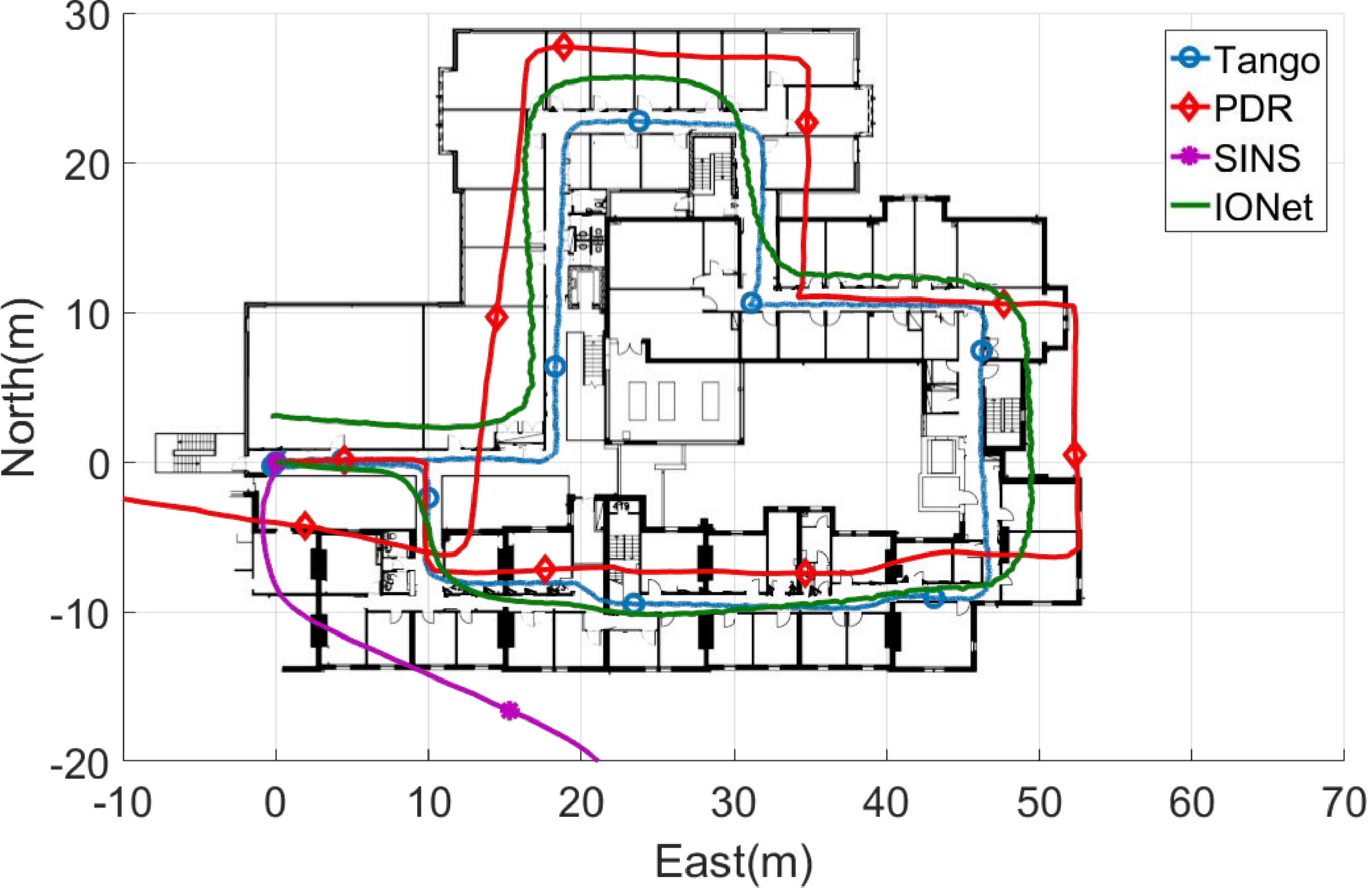}
        	\caption{\label{fig:floor4_hand_phones} Handheld}
        \end{subfigure}
        \begin{subfigure}[t]{0.3\textwidth}
        	\includegraphics[width=\textwidth]{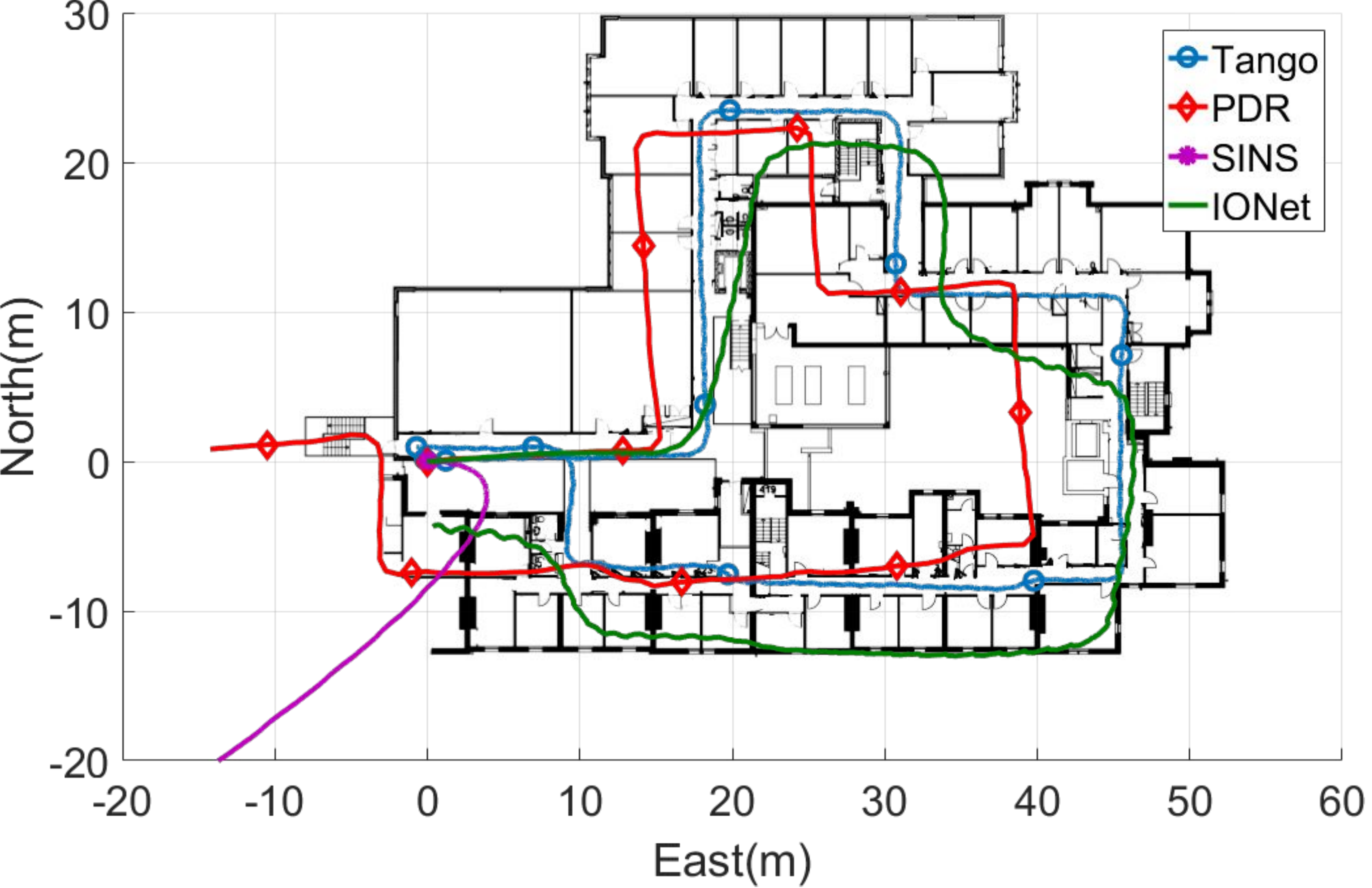}
        	\caption{\label{fig:floor4_pocket_phones} In Pocket}
        \end{subfigure}
        \begin{subfigure}[t]{0.3\textwidth}
        	\includegraphics[width=\textwidth]{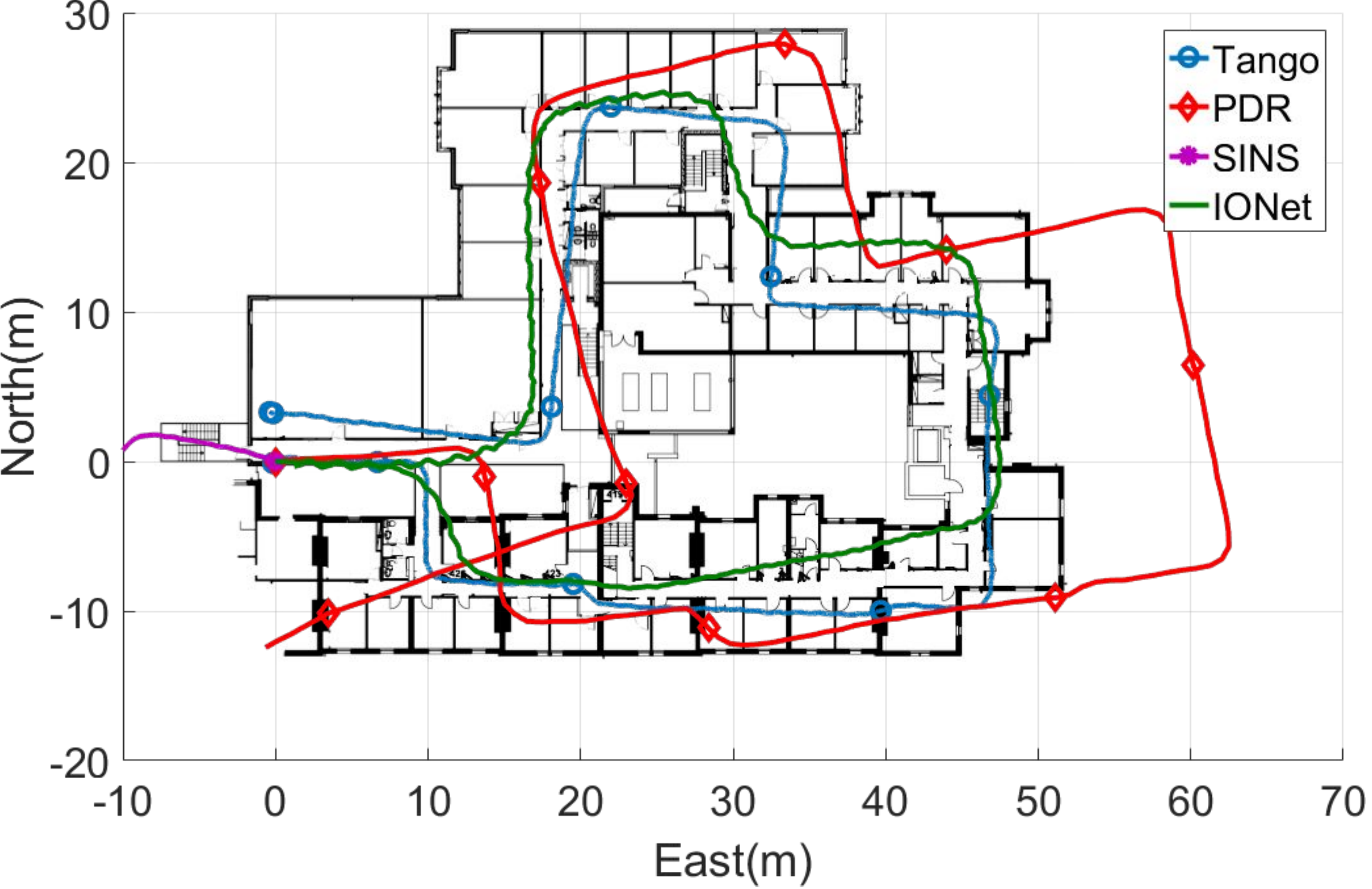}
        	\caption{\label{fig:floor4_handbag_phones} In Handbag}
        \end{subfigure}
        \caption{\label{fig:floor4} Trajectories on Floor B}
    \end{figure*}

	In the physical model, orientation transformations impact all subsequent outputs. We adopt Long Short-Term Memory (LSTM) to handle the exploding and vanishing problems of vanilla RNN, as it has a much better ability to exploit the long-term dependencies \cite{Greff2016}. In addition, as both previous and future frames are crucial in updating the current frame a bidirectional architecture is adopted to exploit dynamic context.
      
	Equation (\ref{eq:polar_vector}) shows that modeling the final polar vector requires modeling some intermediate latent variables, e.g. initial velocity and gravity. Therefore, to build up higher representation of IMU data, it is reasonable to stack 2-layer LSTMs on top of each other, with the output sequences of the first layer supplying the input sequences of the second layer. The second LSTM outputs one polar vector to represent the transformation relation in the processed sequence. Each layer has 96 hidden nodes. To increase the output data rate of polar vectors and locations, IMU measurements are divided into independent windows with a stride of 10 frames (0.1s).
    
	The optimal parameter $\mathbf{\theta}^*$ inside the proposed deep RNN architecture can be recovered by minimizing a loss function on the training dataset $\mathbf{D}=(\mathbf{X},\mathbf{Y})$.   
    \begin{equation}
    	\theta^*= \operatorname*{arg\, min}_{\theta} \ell(f_{\theta}(\mathbf{X}),\mathbf{Y})
    \end{equation}

	The loss function is defined as the sum of Euclidean distances between the ground truth $(\Delta \tilde{l}, \Delta \tilde{\psi})$ and estimated value $(\Delta l, \Delta \psi)$. 
	\begin{equation}
    	\ell=\sum \|\Delta \tilde{l} - \Delta l \|_2^2+\kappa \|\Delta \tilde{\psi} - \Delta \psi \|_2^2
    \end{equation}
	where $\kappa$ is a factor to regulate the weights of $\Delta l$ and $\Delta \psi$.

	\begin{figure}
    	\centering
        \includegraphics[width=0.5\textwidth]{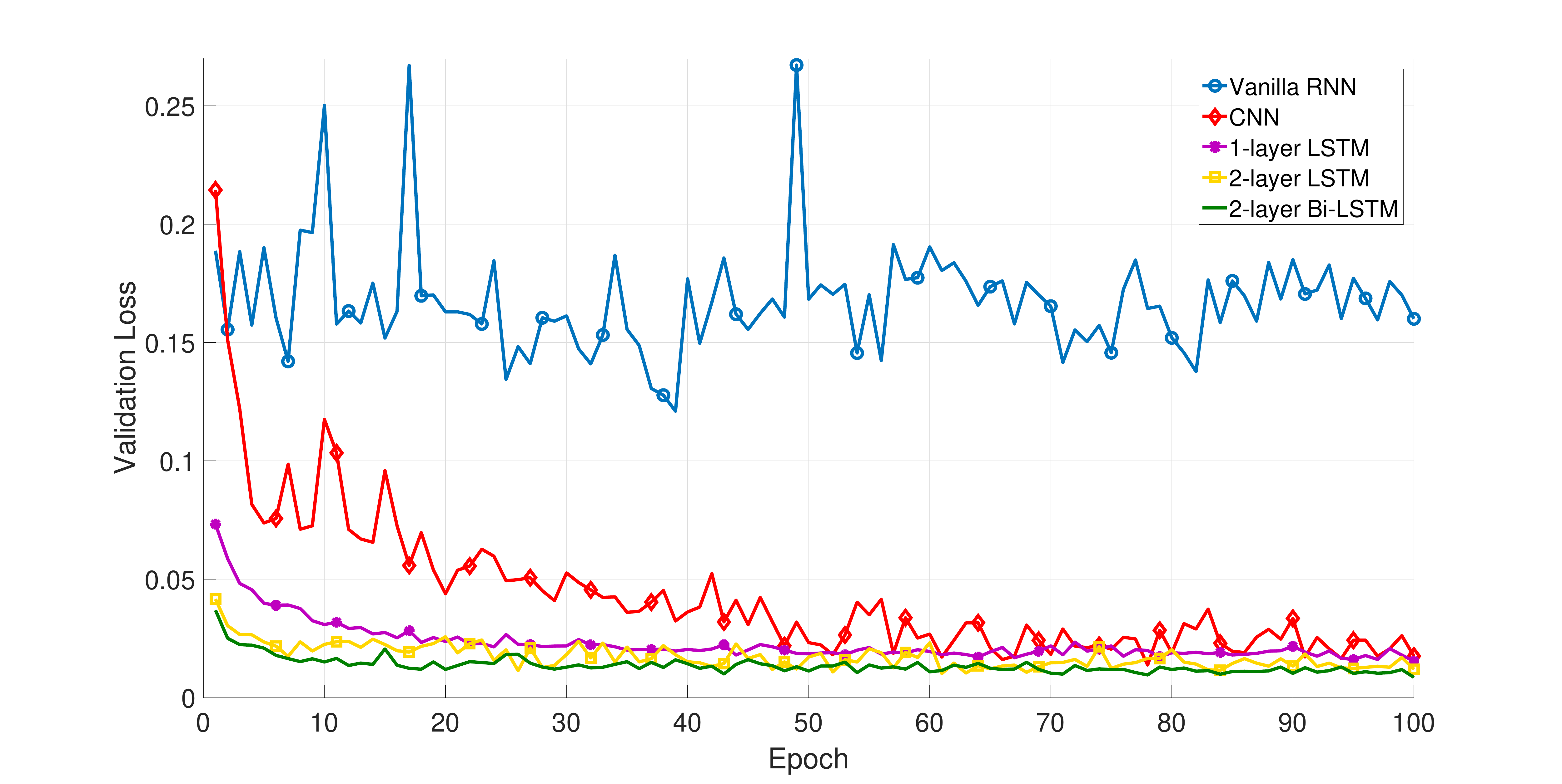}
        \caption{\label{fig:loss}Loss of adopting various frameworks}
    \end{figure}

	 \begin{figure*}
    	\centering
        \begin{subfigure}[t]{0.32\textwidth}
        	\includegraphics[width=\textwidth]{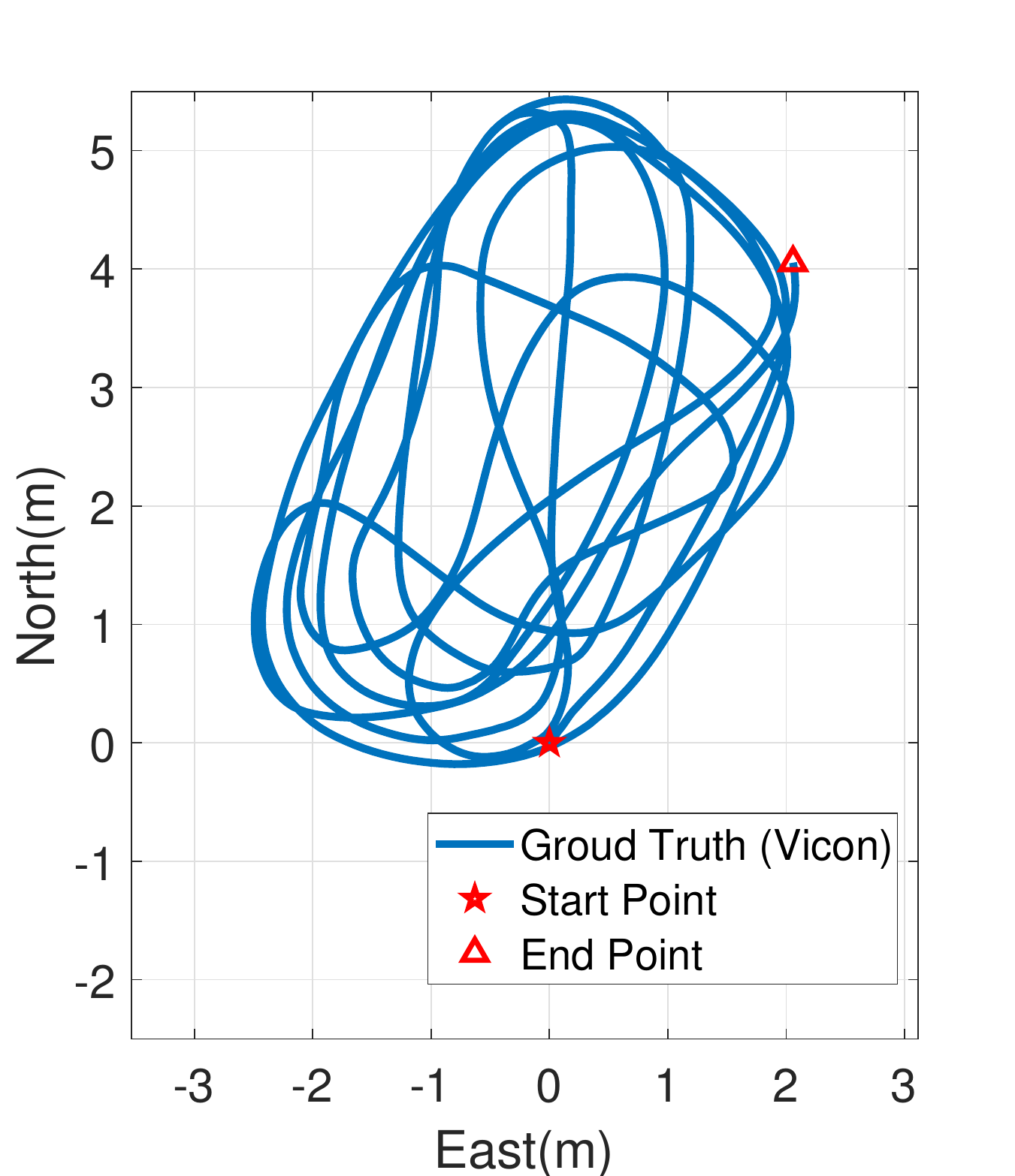}
        	\caption{\label{fig:traj_vicon} Ground Truth}
        \end{subfigure}
        \begin{subfigure}[t]{0.32\textwidth}
        	\includegraphics[width=\textwidth]{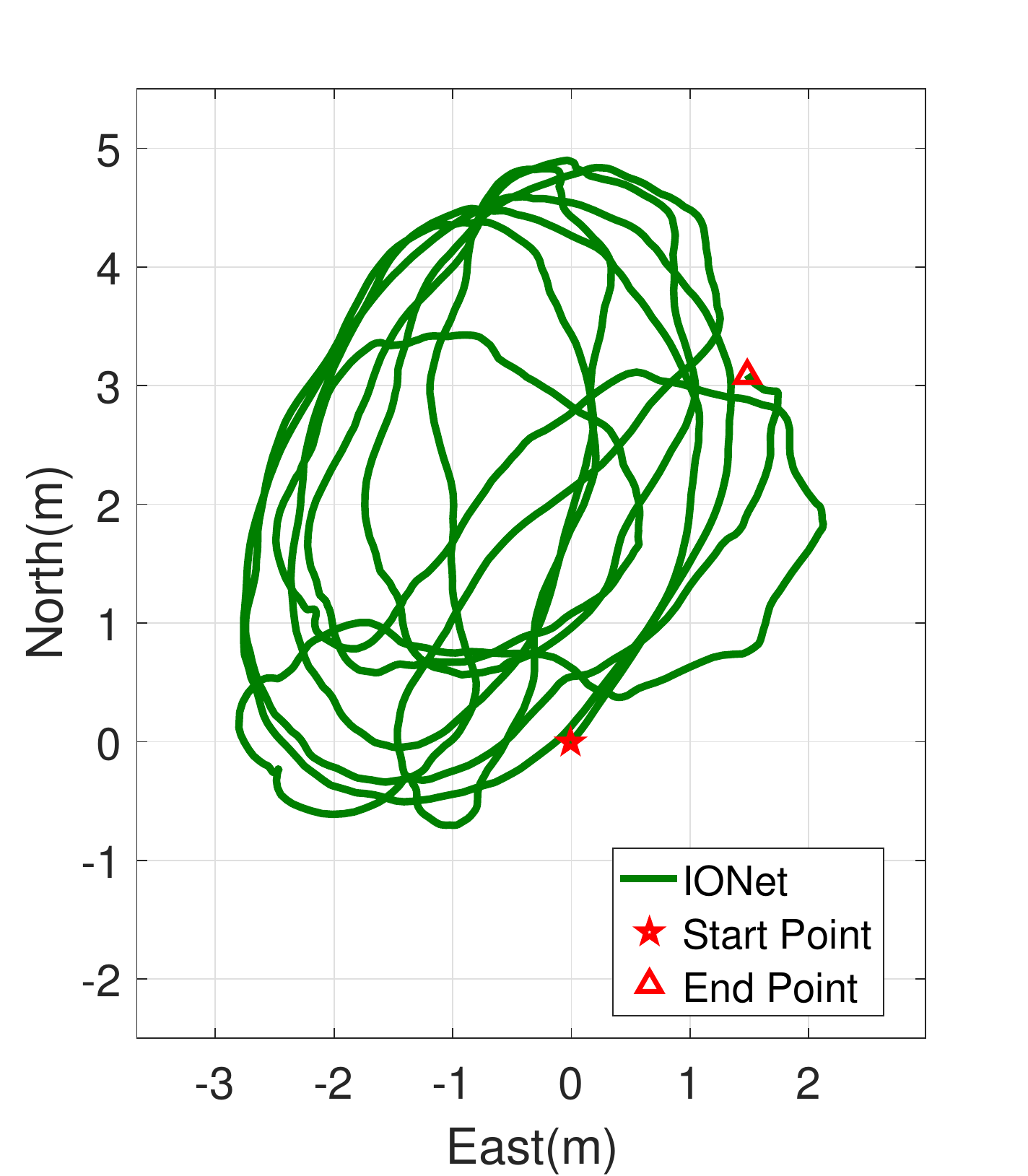}
        	\caption{\label{fig:traj_ionet} IONet}
        \end{subfigure}
        \begin{subfigure}[t]{0.32\textwidth}
        	\includegraphics[width=\textwidth]{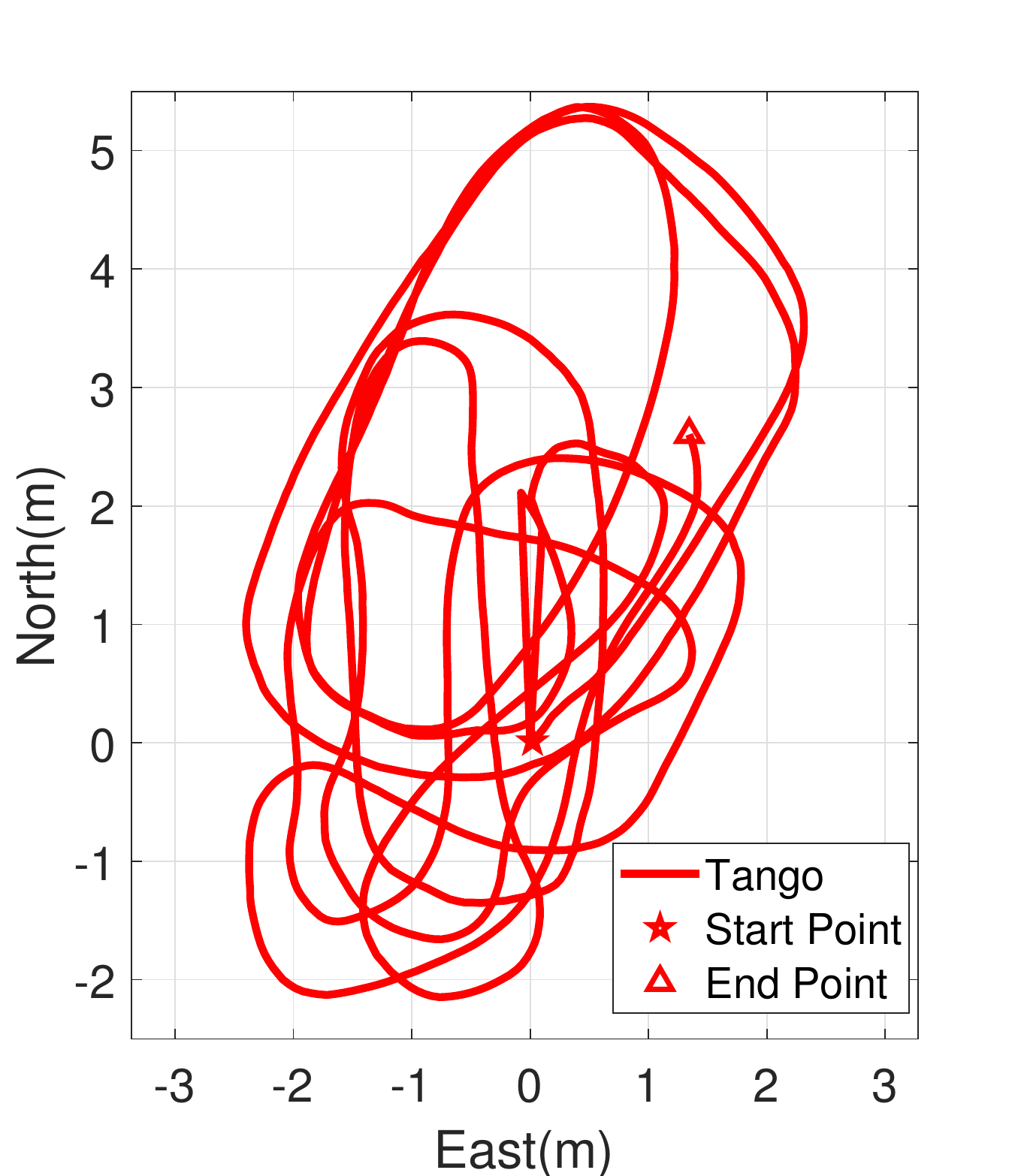}
        	\caption{\label{fig:traj_tango} Tango}
        \end{subfigure}
        \caption{\label{fig:traj_trolley} Trolley tracking trajectories of (a) Ground Truth (b) IONet (c) Tango}
    \end{figure*} 
 
\section{Experiments}

\subsection{Training Details}

	\textbf{Dataset}: There are no public datasets for indoor localization using phone-based IMU. We collected data ourselves with pedestrian walking inside a room installed with an optical motion capture system (Vicon) \cite{Vicon2017}, providing very high-precise full pose reference (0.01m for location, 0.1 degree for orientation) for our experimental device. The training dataset consists of IMU data from the phone in different attachments, e.g. hand-held, in pocket, in handbag, on trolley each for 2 hours, collected by a very common consumer phone iPhone 7Plus. Note that all of our training data was collected by User 1 carrying iPhone 7. To test our model's ability to generalize, we invited 3 new participants and evaluated on another two phones, i.e. iPhone 6 and iPhone 5. 
    
    \textbf{Training}: We implemented our model on the public available TensorFlow framework, and ran training process on a NVIDIA TITAN X GPU. During training, we used Adam, a first-order gradient-based optimizer \cite{Kingma2014} with a learning rate of 0.0015. The training converges typically after 100 iterations. To prevent our neural networks from overfitting, we gathered data with abundant moving characteristics, and adopted Dropout \cite{Srivastava2014} in each LSTM layer, randomly dropping 25\% units from neural networks during training. This method significantly reduces overfitting, and proves to perform well on new users, devices and environments. 

	\textbf{Comparison with Other DNN Frameworks}: To evaluate our assumption of adopting a 2-layer Bidirectional LSTM for polar vector regression, we compare its validation results with various other DNN frameworks, including frameworks using vanilla RNN, vanilla Convolution Neural Network, 1-layer LSTM and 2-layer LSTM without Bi-direction. The training data are from all attachments. Figure \ref{fig:loss} shows their validation loss lines. Our proposed framework with 2-layer Bi-LSTM descends more steeply, and stays lower and more smoothly during the training than all other neural networks, supporting our assumption, while vanilla RNN is stuck in vanishing problems, and CNN doesn't seem to capture temporary dependencies well. 
    
    \textbf{Testing:} We also found that a separate training on every attachment shows better performance in prediction than training jointly, hence we implemented the prediction model of 2-layer Bi-LSTM trained on separate attachments in our following experiments. In a practical deployment, existing techniques can be adopted to recognize different attachments from pure IMU measurements \cite{Brajdic2013}, providing the ability to dynamically switch between trained models.

	\textbf{Baselines}: Two traditional methods are selected as baselines, pedestrian dead reckoning (PDR) and strapdown inertial navigation system (SINS) mechanism \cite{Savage1998}, to compare with our prediction results. PDR algorithms are seldom made open-sourced, especially a robust PDR used in different attachments, so we implement code ourselves according to \cite{Brajdic2013} for step detection and \cite{Xiao2014} for heading and step length estimation.

\subsection{Tests Involving Multiple Users and Devices}
	
    A series of experiments were conducted inside a large room with new users and phones to show our neural network's ability to generalize. Vicon system provides highly accurate reference to measure the location errors.

\begin{figure}
    	\centering
        \begin{subfigure}[t]{0.23\textwidth}
        	\includegraphics[width=\textwidth]{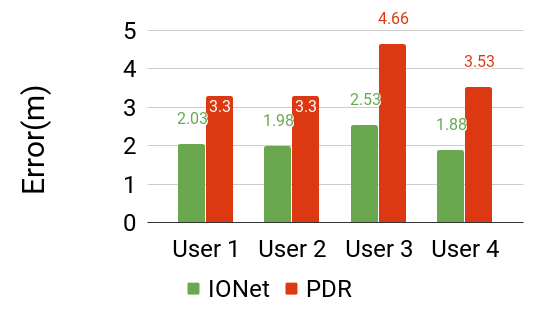}
        	\caption{\label{fig:error_users} Multiple Users}
        \end{subfigure}
        \begin{subfigure}[t]{0.23\textwidth}
        	\includegraphics[width=\textwidth]{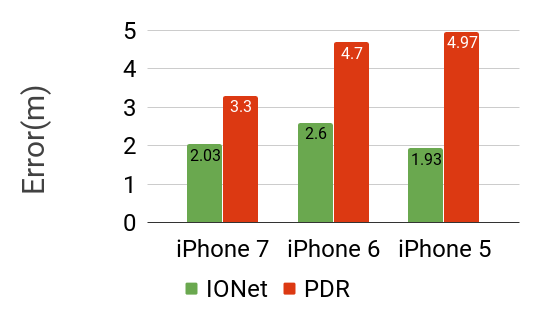}
        	\caption{\label{fig:error_phones} Multiple Phones}
        \end{subfigure}
        \caption{\label{fig:small_error}Maximum position error within 90\% test time}
    \end{figure}

    The first group of tests include four participants, walking randomly for two minutes with the phone in different attachments, e.g. in hand, pocket and handbag respectively, covering everyday behaviors. Our training dataset doesn't contain data from three of these participants. The performance of our model is measured as error cumulative distribution function (CDF) against Vicon ground truth and compared with conventional PDR and SINS. Figure \ref{fig:multi_users} illustrates that our proposed approach outperforms the competing methods in every attachment. If raw data is directly triply integrated by SINS, its error propagates exponentially. The maximum error of our IOnet stayed around 2 meter within 90\% testing time, seeing 30\%- 40\% improvement compared with traditional PDR in Figure \ref{fig:error_users}.

	Another group of experiments is to test the performance across different devices, shown in Figure \ref{fig:multi_phones}. We choose another two very common consumer phones, iPhone 6 and iPhone 5, whose IMU sensors, InvenSense MP67B and ST L3G4200DH, are quite distinct from our training device, iPhone 7 (IMU: InvenSense ICM-20600). Although intrinsic properties of IMU influence the quality of inertial measurements, our neural network shows good robustness.

\subsection{Large-scale Indoor Localization}

	Here, we apply our model on more challenging indoor localization experiment to present its performance in a new environment. Our model \textit{without training outside} Vicon room, is directly applied to six large-scale experiments conducted on two floors of an office building. The new scenarios contained long straight lines and slopes, which were not contained in the training dataset. Lacking the high precise reference from Vicon, we take Google Tango Tablet \cite{Tango}, a famous visual-inertial device, as pseudo ground truth.
  
 	\begin{figure}
    	\centering
        \begin{subfigure}[t]{0.23\textwidth}
        	\includegraphics[width=\textwidth]{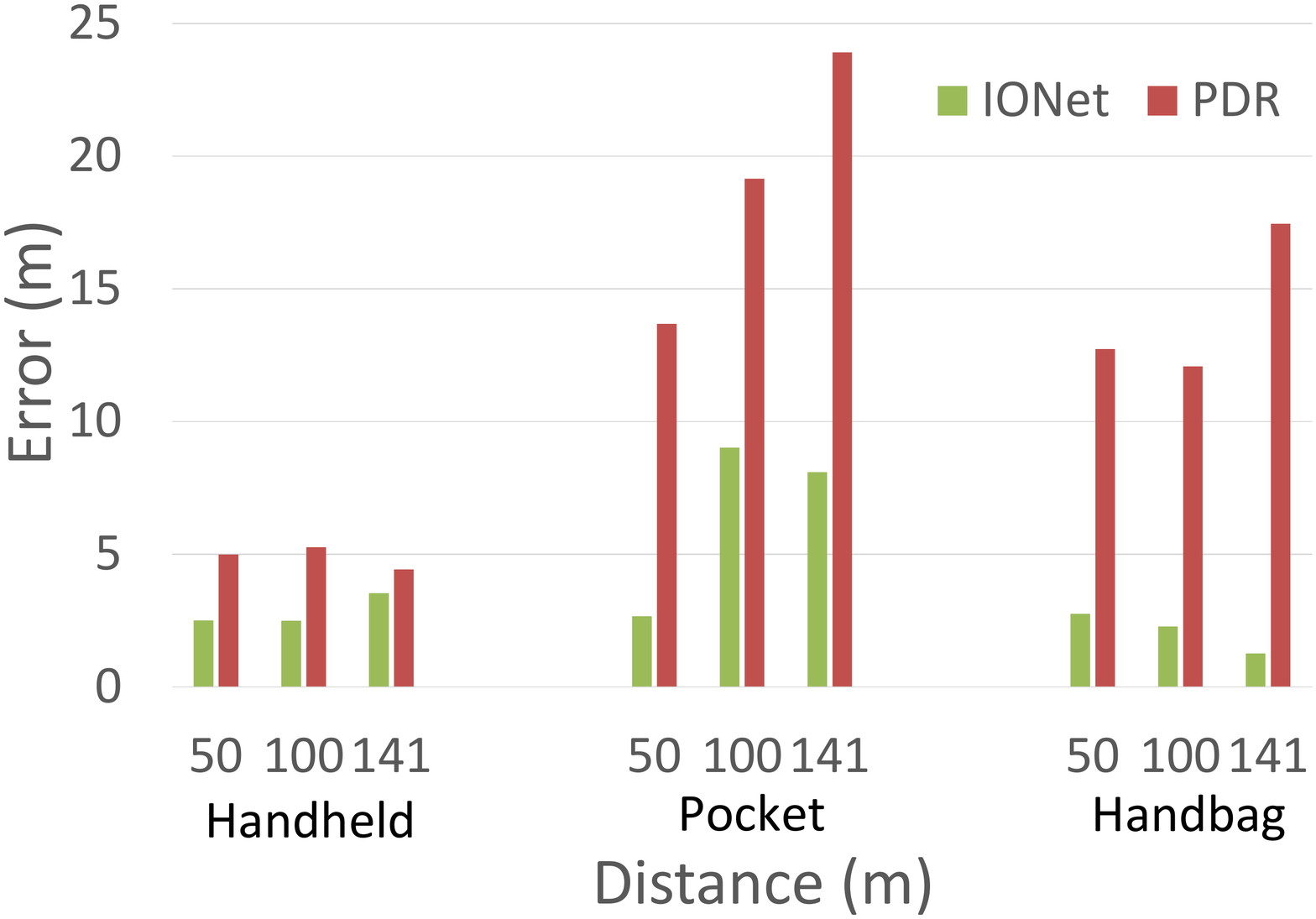}
        	\caption{\label{fig:floor1_err} Floor A}
        \end{subfigure}
        \begin{subfigure}[t]{0.23\textwidth}
        	\includegraphics[width=\textwidth]{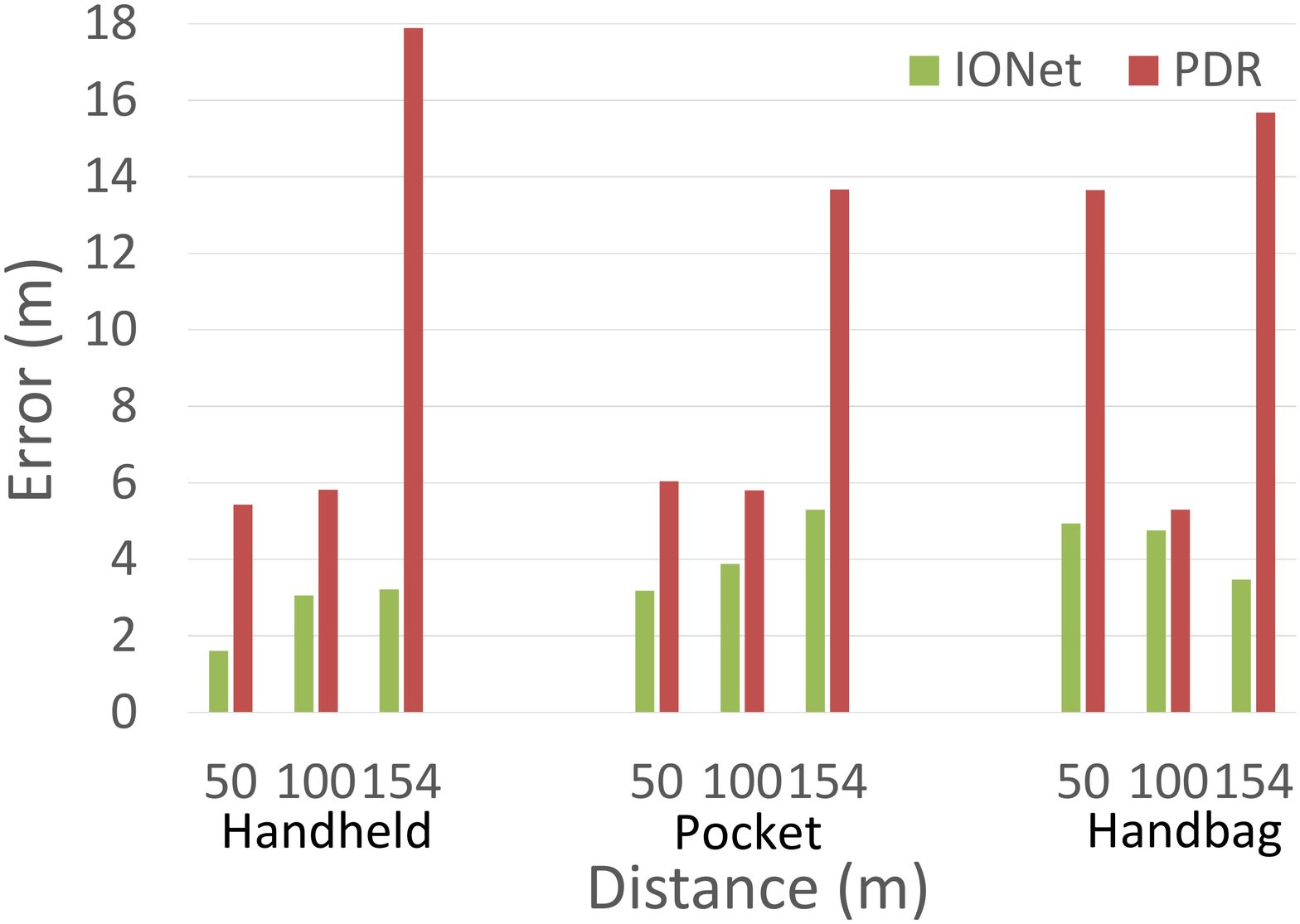}
        	\caption{\label{fig:floor4_err} Floor B}
        \end{subfigure}
        \caption{\label{fig:large_error} Position error in large-scale indoor localization}
    \end{figure}

    The floor maps are illustrated in Figure \ref{fig:floor1} (about 1650 $m^2$ size) and Figure \ref{fig:floor4} (about 2475 $m^2$). Participants walked normally along corridors with the phone in three attachments respectively. The predicted trajectories from our proposal are closer to Tango trajectories, compared with the two other approaches in Figure \ref{fig:floor1} and \ref{fig:floor4}. The continuously propagating error of SINS mechanism caused trajectory drifts that grow exponentially with time. Impacted by wrong step detection or inaccurate step stride and heading estimation, PDR accuracy is limited. We calculate absolute position error against pseudo ground truth from Tango at a distance of 50m, 100m and the end point in Figure \ref{fig:large_error}. Our IONet shows competitive performance over traditional PDR and has the advantage of generating continuous trajectory at 10 Hz, though its heading attitude deviates from true values occasionally.

\subsection{Trolley Tracking}
	
    We consider a more general problem without periodic motion, which is hard for traditional step-based PDR or SINS on limited quality IMU. Tracking wheel-based motion, such as a shopping trolley/cart, robot or baby-stroller is highly challenging and hence under-explored. Current approaches to track wheeled objects are mainly based on visual odometry or visual-inertial odometry (VIO) \cite{Li2013b,Bloesch2015}. They won't work when the device is occluded or operating in low light environments, such as placed in a bag. Moreover, their high energy- and computation-consumption also constrain further application. Here, we apply our model on a trolley tracking problem \textit{using only inertial sensors}. Due to a lack of comparable technique, our proposal is compared with the state-of-art visual-inertial odometry Tango.
    
    \begin{figure}
    	\centering
        \includegraphics[width=0.45\textwidth]{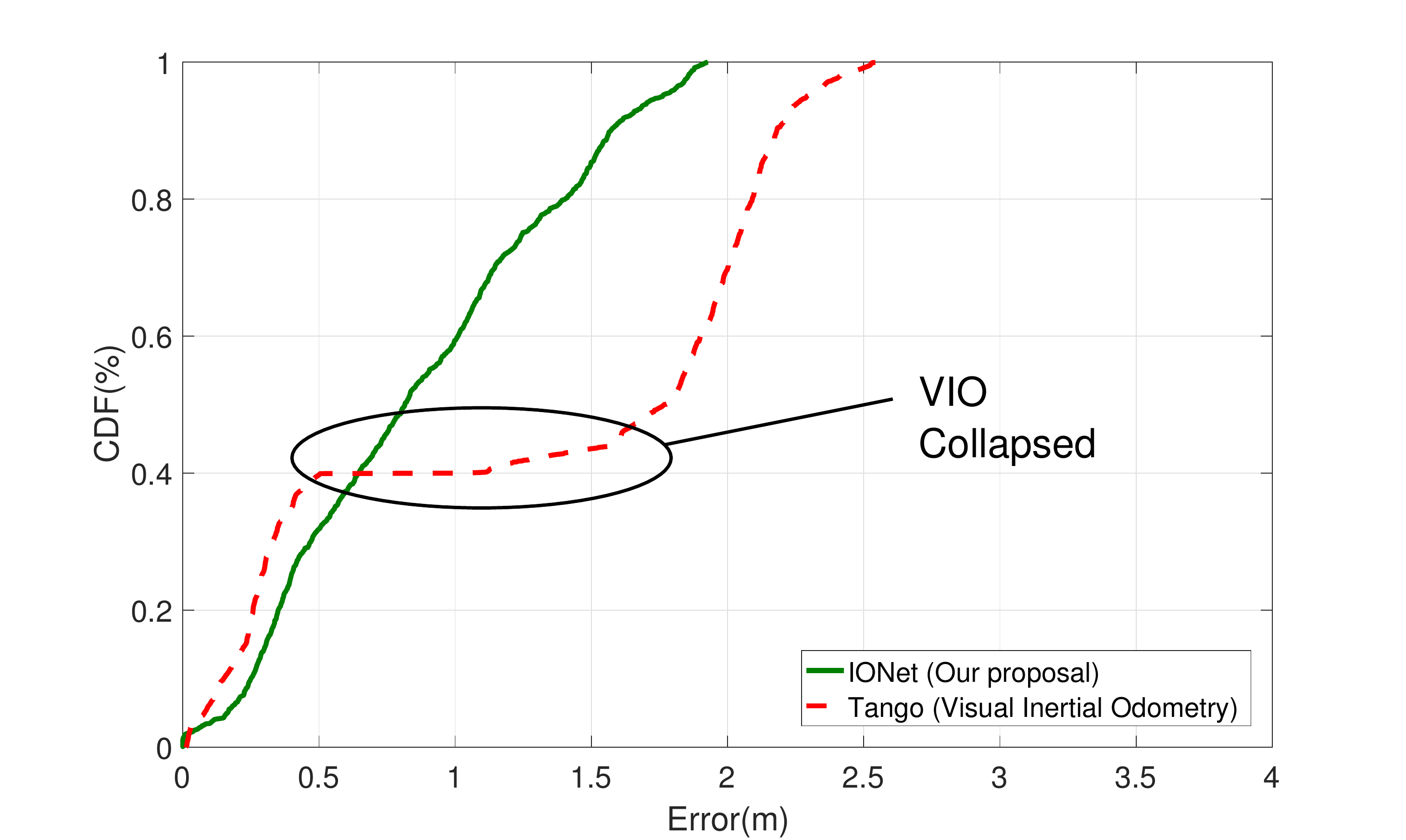}
        \caption{\label{fig:trolley_cdf}CDF of Trolley Tracking}
    \end{figure}  
    
    Our experiment devices, namely an iPhone 7 and the Google Tango are attached on a trolley, pushed by a participant. Detailed experiment setup and results could be found in supplementary video \footnote{Video is available at: https://youtu.be/pr5tR6Wz-zs}. High-precision motion reference was provided by Vicon. From the trajectories from Vicon, our IONet and Tango in Figure \ref{fig:traj_trolley}, our proposed approach shows almost the same accuracy as Tango, and even better robustness, because our pure inertial approach suffers less from environmental factors. With the help of visual features, VIO (Tango) can constrain error drift by fusing visual transformations and the inertial system, but it will collapse when capturing wrong features or no features, especially in open spaces. This happened in our experiment, shown in Figure \ref{fig:trolley_cdf}. Although VIO can recover from the collapse, it still left a large distance error.

\section{Conclusion and Future Work}

We presented a new neural network framework to learn inertial odometry directly from IMU raw data. Our model is derived from Newtonian mechanics and formulated as a sequential learning problem using deep recurrent neural networks to recover motion characteristics. The performance of IONet is evaluated through extensive experiments including tests involving multiple users/devices, large-scale indoor localization and trolley tracking. It outperforms both traditional step-based PDR and SINS mechanisms. We believe our work lays foundations for a more accurate and reliable indoor navigation system fusing multiple sources. 

A number of research challenges lie ahead: 1)  The performance of our proposed IONet degraded when the input IMU measurements are corrupted with a high bias value, because the training and testing data are not in the same domains. 2) Challenging motions and distinct walking habits of new users can influence the effectiveness of our proposed approach. 3) Our current model is trained and tested on different attachments separately. The challenge becomes how to jointly train all the attachments on one model and improve its robustness. 4) Combining the sequence-based physical model with the proposed neural network model would be an interesting point to investigate in future.

Therefore, our future work includes extending our work to learn transferable features without the limitations of measurement units, users and environments, realized on all platforms including mobile devices and robots. 

\section{ Acknowledgments}
We would like to thank the anonymous reviewers for their valuable comments. We also thank our colleagues, Dr. Sen Wang, Dr. Stepheno Rosa, Shuyu Lin, Linhai Xie, Bo Yang, Jiarui Gan, Zhihua Wang, Ronald Clark, Zihang Lai, for their help, and Prof. Hongkai Wen at University of Warwick for useful discussions and the usage of GPU computational resources.

\bibliographystyle{aaai}
\bibliography{Mendeley.bib}

\end{document}